\documentclass[man,floatsintext,longtable]{apa7}

\usepackage{lipsum}

\usepackage[american]{babel}

\usepackage{csquotes}
\usepackage[style=apa,sortcites=true,sorting=nyt,backend=biber]{biblatex}
\DeclareLanguageMapping{american}{american-apa}
\usepackage{CJKutf8}

\usepackage{amsmath}
\usepackage[shortlabels]{enumitem}
\usepackage{bbm}
\usepackage{indentfirst}
\usepackage{xcolor}
\usepackage{booktabs}
\usepackage{multirow}
\usepackage{pbox}
\usepackage{float}

\usepackage{tabularx,ragged2e}

\usepackage{balance}
\usepackage{algorithm}
\usepackage{algpseudocode}

\usepackage{subfigure}

\usepackage{url}
\usepackage{stfloats}

\usepackage{bm}

\usepackage[figuresright]{rotating}
\usepackage{longtable}

\usepackage{color}
\usepackage{xcolor}
\usepackage{stix}

\title{Discovering Semantic Structures in Psychological Scales: A Response-Free Front-End to Efficient Simplification}
\shorttitle{A Response-Free Front-End to Efficient Scale Simplification}

\author{
Bo Wang\textsuperscript{1,\textdagger},
Yuxuan Zhang\textsuperscript{2,\textdagger},
Yueqin Hu\textsuperscript{3,4},
Hanchao Hou\textsuperscript{2},
Kaiping Peng\textsuperscript{1},
Shiguang Ni\textsuperscript{2,*}
}

\affiliation{
\textsuperscript{1}{Department of Psychological and Cognitive Sciences, Tsinghua University, Beijing, China} \\
\textsuperscript{2}{Shenzhen International Graduate School, Tsinghua University, Shenzhen, China.}\\
\textsuperscript{3}{Faculty of Psychology, Beijing Normal University, Beijing, China.}\\
\textsuperscript{4}{Beijing Key Laboratory of Applied Experimental Psychology, National Demonstration Center for Experimental Psychology Education, Faculty of Psychology, Beijing Normal University, Beijing, China.}
}

\leftheader{Weiss}

\abstract{
Psychological scale refinement traditionally relies on response-based methods such as factor analysis, item response theory, and network psychometrics to uncover latent constructs and optimize item composition. Although methodologically rigorous, these approaches require large respondent samples and may be limited by data availability, cross-cultural comparability, and replication constraints. 
Recent advances in natural language processing suggest that the wording of questionnaire items contains construct-relevant semantic organization, offering a complementary response-free perspective. 
We introduce a topic-modeling framework that operationalizes item-level semantic structure for systematic scale simplification. Items are encoded using contextual sentence embeddings and grouped through density-based clustering to identify semantic topics without predefining their number. 
Within each cluster, class-based term weighting identifies representative keywords that summarize each topic and support an interpretable semantic pre-structure of the item pool. Semantically related clusters are hierarchically merged, and representative items are selected using structure-aware membership criteria within an integrated reduction pipeline. We evaluated the framework on three widely used instruments: DASS, IPIP, and EPOCH, covering negative affect, personality traits, and well-being. Results showed that the identified semantic topics often formed factor-like groupings aligned with established constructs.
Representative items selected through semantic membership probabilities reduced item counts by 60.5\% on average while maintaining acceptable psychometric properties and preserving factor correlation patterns.
These findings support semantic topic-structure modeling as an interpretable front-end for theory-consistent scale construction and simplification, guiding rather than replacing subsequent psychometric validation.
An integrated visualization-supported tool further enables one-click semantic analysis and structured item selection.
}


\authornote{
  \addORCIDlink{Bo Wang}{0000-0001-7587-5141}\\
  \addORCIDlink{Yuxuan Zhang}{0000-0002-6179-4313}\\
  \addORCIDlink{Yueqin Hu}{0000-0002-5533-9277}\\
  \addORCIDlink{Hanchao Hou}{0000-0002-5019-3010}\\
  \addORCIDlink{Kaiping Peng}{0000-0001-8515-3409}\\
  \addORCIDlink{Shiguang Ni}{0000-0002-4303-7386}\\

 \noindent \textsuperscript{\textdagger}These authors contributed equally: Bo Wang, Yuxuan Zhang. 
  
\noindent  *Correspondence concerning this article should be addressed to: Shiguang Ni, Shenzhen International Graduate School, Tsinghua University. E-mail: ni.shiguang@sz.tsinghua.edu.cn}

\begin{document}
\maketitle

The psychological questionnaire is a central tool in psychometric assessment, translating latent psychological constructs into observable, interpretable scores. Yet developing, refining, and administering such instruments is often costly: items must be iteratively drafted, piloted, and revised, and longer forms increase respondent burden, attrition, and careless responding. Accordingly, there is sustained demand for short-form scales that remain psychometrically defensible while improving feasibility, participant experience, and downstream analysis.

\subsection{Psychometric Methods for Scale Simplification}
The most commonly used approaches to scale simplification are traditional psychometric methods. Classic Test Theory (CTT)-based procedures typically rely on item-total correlations and internal consistency indices to identify redundant or weak items \parencite{stanton2002issues, patrick2019many}. Similarly, Principal Component Analysis (PCA) and Factor Analysis (FA) models, as both Exploratory and Confirmatory Factor Analyses (EFA, CFA), are widely used to identify a low-dimensional structure in item responses and to retain items that load strongly on the resulting components or theoretically meaningful factors \parencite{tavakol2020factor,wise2023identifying}. Item Response Theory (IRT) further refines this process by modeling item discrimination and difficulty parameters, often yielding highly efficient short forms when adequate response data are available \parencite{leipoldt2018refining,o2018practical,morrison2022optimized}.

More recently, network-based approaches, such as Exploratory Graph Analysis (EGA) \parencite{golino2020investigating}, reconceptualize questionnaire items as nodes connected by partial correlations, enabling the identification of communities or central items through graph topology rather than latent variables. It offers an interpretable structure by modeling latent relational patterns among items. The EGA and its extensions \parencite{golino2020investigating,jamison2024metric} have been widely applied to validate and shorten scales across clinical and personality domains \parencite{golino2020investigating,maertens2024misinformation,portoghese2024network,fuchshuber2025uncovering,abdelrahman2025psychometric}.

Taken together, these approaches either select items that maximize empirical reliability or information, such as CTT, PCA, and IRT, or model interpretable structures from observed response patterns, such as FA and EGA. In both cases, item selection is grounded in response data and established psychometric evidence; however, this also means that semantic redundancy and conceptual overlap embedded in item wording are usually examined only indirectly, after data have been collected.

\subsection{Conventional Machine Learning-Assisted Simplification}
Parallel to these psychometric traditions, machine learning-assisted approaches have been increasingly explored for questionnaire simplification. 
One line of work uses supervised feature-selection models, such as logistic regression, support vector machines, and random forests, to identify minimal item subsets that approximate full-scale scores or external outcomes \parencite{yu2024simplification,xu2025machine,jiang2023developing,lin2022using}. Another line uses evolutionary or optimization-based strategies, including genetic algorithms and ant colony optimization, to search for item combinations that maximize predefined psychometric criteria under length constraints \parencite{passarelli2024short,cao2025method}. Hybrid pipelines further combine machine learning with classical psychometric checks, such as response-based clustering or variable grouping followed by CFA or IRT confirmation \parencite{cai2025machine,lee2023simplification}.
Although these approaches can improve predictive efficiency and automate item selection, they typically remain dependent on respondent data, labeled outcomes, or repeated evaluation against response-based criteria. Moreover, because item texts are rarely modeled as semantic objects in their own right, these methods usually do not provide an explicit, interpretable account of semantic redundancy or conceptual substructure within the item pool.

\subsection{Introducing a Topic-modeling Framework for Scale Simplification}

Natural language processing (NLP) has increasingly been integrated into psychological research, evolving from dictionary-based word matching approaches (e.g., LIWC, \parencite{tausczik2010psychological}) to open-vocabulary and embedding-based models capable of capturing latent psychological patterns in natural language. Large-scale analyses of social media text have demonstrated that linguistic representations can reliably infer personality traits and other stable individual differences \parencite{schwartz2013personality,preotiuc2016studying}, while clinical applications have shown that language patterns can signal psychological risk and distress \parencite{milne2016clpsych,coppersmith2018natural}. More recently, contextual embedding frameworks have begun to link psychometric theory with semantic modeling, operationalizing psychologically grounded dimensions directly from text \parencite{araque2020moralstrength,atari2023construct,chen2024surveying,jocelyn2025large}.
These developments suggest that linguistic representations can approximate psychologically meaningful constructs, raising the possibility that the semantic structure of questionnaire items themselves may encode latent measurement organization.

Several lines of research suggest that item semantics contain measurable structural information. At the word level, \Textcite{rosenbusch2020semantic} introduced the Semantic Scale Network, using latent semantic analysis (LSA)-based representations to quantify semantic overlap between scales without participant data. Their work highlighted the importance of systematic lexical comparison for detecting redundancy, though analyses were conducted at the scale (document) level rather than modeling latent item-level structure. Moving beyond token-level representations, \Textcite{herderich2024computational} and \Textcite{huang2025embedding} employed sentence embeddings and clustering techniques to explore semantic organization among psychological texts and questionnaire items, demonstrating that contextual embeddings can reveal meaningful grouping patterns. Complementing these structural explorations, \Textcite{ravenda2025rethinking} provided empirical evidence that semantic similarity among items strongly correlates with empirical score correlations, suggesting that semantic matrices approximate response-based factor structure. Finally, \Textcite{jung2025transformer} applied an embedding-based clustering with K-Means to support item reduction, illustrating the practical potential of semantic grouping for scale compression.

To conclude, past studies establish that (a) lexical overlap can signal redundancy, (b) contextual embeddings can uncover semantic organization, and (c) semantic similarity approximates empirical correlation. However, they either operate at the scale level, rely on predefined cluster numbers, focus on pairwise similarity, or treat clustering as a heuristic step rather than a structured, inspectable reduction framework. In particular, partition-based clustering methods (e.g., \parencite{jung2025transformer}) require the number of clusters to be specified a priori and select representative items solely based on geometric proximity, which may obscure density structure or blur the semantically adjacent factors.

Building on these developments, we propose a topic-modeling framework that treats semantic structure as a response-free front-end for scale construction and simplification. Rather than limiting analysis to word-level matching, pairwise similarity, or fixed clustering schemes, our framework explicitly models semantic structure at the item level using density-based clustering, followed by topic representation and controlled merging procedures. By integrating contextual sentence embeddings with cluster-level topic modeling, we capture both global semantic geometry and interpretable keyword structure, enabling systematic representative item selection within an operational pipeline.

\begin{figure}[ht]
    \caption{Overall Framework}
    \includegraphics[width=1\linewidth]{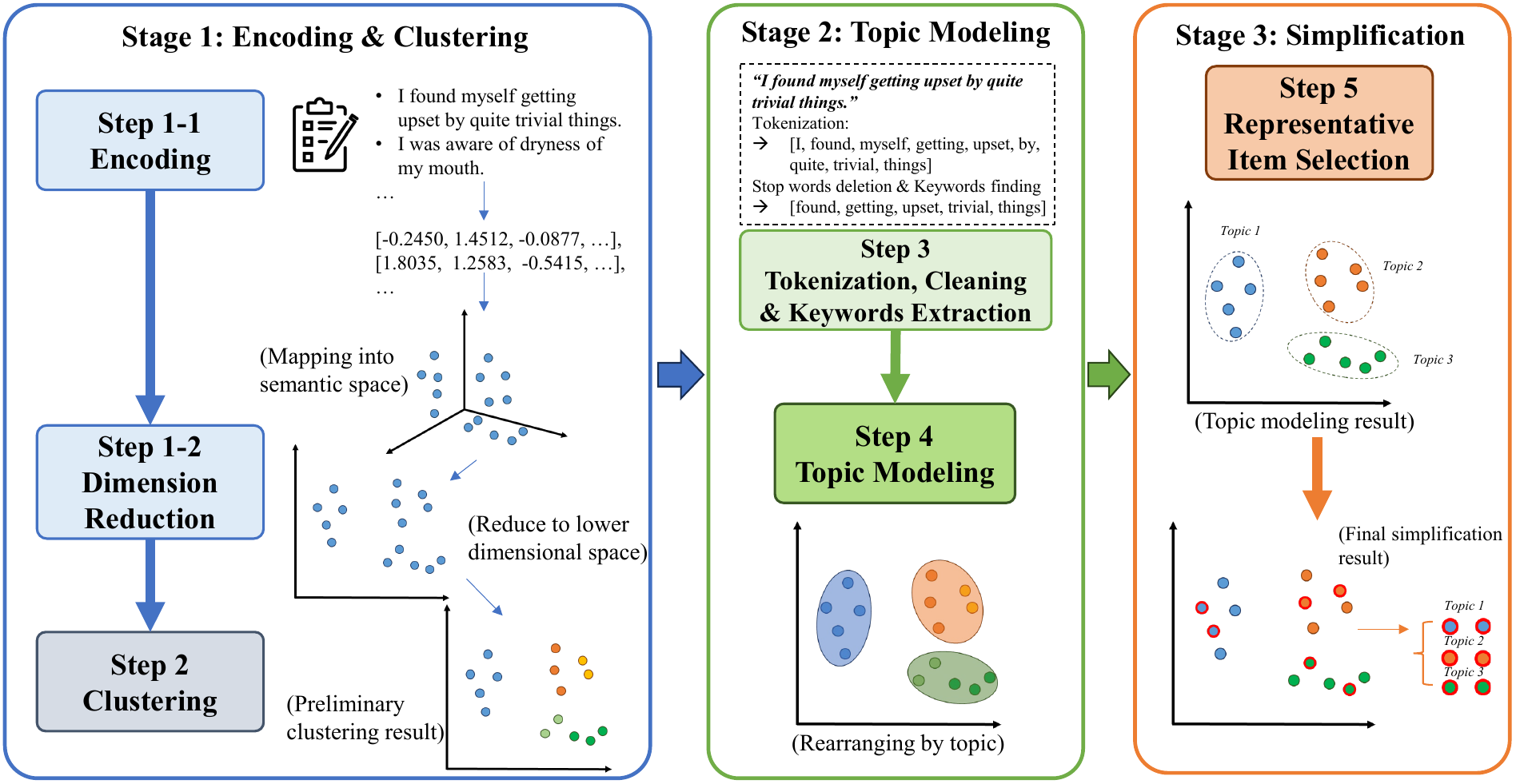}
    \label{fig:intro-framework}
\end{figure}

Specifically, as shown in Figure \ref{fig:intro-framework}, our framework first encodes each item into a contextual semantic embedding space (step 1-1), where items with similar meanings are positioned closer together. We then apply dimensionality reduction (step 1-2) and density-based clustering (step 2) to identify coherent semantic groupings. After that, the original item textual data are also processed (step 3) and jointly formalize the final topic clusters (step 4) by a structured merging of semantically adjacent clusters, where high-frequency and distinctive terms (i.e., keywords) are extracted to characterize the thematic core, yielding interpretable semantic topics. Finally, representative items are selected using membership probability within each factor (step 5), which is analogous in spirit to prioritizing high-loading items in traditional factor-analytic reduction, but derived from semantic structure rather than response covariance.

Compared to prior machine learning–assisted simplification approaches, our framework formalizes semantic structure as an inspectable, response-free approximation of item-level content organization that guides and constrains subsequent empirical evaluation. By combining contextual embeddings, density-based clustering, interpretable topic modeling, and structured representative selection within a unified pipeline, our method extends prior semantic approaches into a comprehensive framework for scale simplification. A detailed comparison with existing approaches is presented in Table \ref{tab:relate-compare}.

\begin{sidewaystable}[ph!]
  \centering
  \caption{Summarization of Scale Simplification Methods}
  	\resizebox{1\linewidth}{!}{
    \begin{tabular}{m{20em}<{\centering}m{7em}<{\centering}m{20em}<{\centering}m{20em}<{\centering}m{15em}}
    \toprule
    Approach Category & Requires Response Data & Interpretability of Structure & Assumes Fixed Dimensionality & Primary Output \\
    \midrule
    Classical psychometrics (CTT/FA/IRT) & Yes   & High (factor loadings, item parameters) & Often yes (theory- or data-driven) & Short form + latent factor model \\
    Network-based psychometrics (EGA, NA) & Yes   & Moderate–High (communities, centrality) & No (data-driven) & Structural representation + candidate items \\
    ML-assisted selection (supervised / optimization-based) & Yes   & Low–Moderate (model-dependent) & Implicitly yes (via objective function) & Predictive short form \\
    Embedding-based clustering with fixed clustering number & No    & Low–Moderate (geometric proximity) & Yes (predefined K) & Compressed item set \\
    Proposed topic-modeling framework & No    & High (semantic topics + labels) & No (density-based) / Optional & Interpretable semantic pre-structure + short form \\
    \bottomrule
    \end{tabular}}%
  \label{tab:relate-compare}%
\end{sidewaystable}%

\subsection{The Current Study}
Based on the existing literature on psychometric and machine learning-assisted approaches to scale simplification, two major limitations were identified. First, most scale-shortening techniques require substantial response datasets to estimate covariance structures, fit models, or optimize selection criteria, which can be costly during early-stage scale development, cross-cultural adaptation, or in data-constrained settings. Second, because item texts are typically treated primarily as indicators rather than as objects of analysis, semantic redundancy and conceptual overlap embedded in wording often remain underexploited. 
Accordingly, this study aims to (1) propose a response-free, unsupervised topic-modeling framework for identifying item-level semantic topic structures within psychological scales, (2) evaluate the proposed method using multiple validated scales spanning distinct construct domains to examine whether these semantic structures align with established theoretical factors and can generate candidate short forms that preserve key psychometric properties after validation, and (3) release an open-source, user-friendly toolkit to support future interpretable and efficient scale refinement for psychological researchers.

\newpage

\section{Methodology}
\subsection{Constructing a Topic-modeling Framework for Scale Simplification}
\subsubsection{Problem Definition}
The proposed framework formulates questionnaire simplification as a response-free semantic structure discovery and representative-item selection problem. Let a psychological scale consist of a finite set of item texts:

\begin{equation}
    \mathcal{X} = \{x_1,x_2,...,x_N\}
\end{equation}
where each $x_i$ is a natural-language statement intended to operationalize a psychological construct.

\paragraph{Inputs}
The sole input to the framework is the item text set $\mathcal{X}$. No respondent data, item scores, or empirical covariance information are required at any stage of item reduction.

\paragraph{Outputs}
The framework produces three complementary outputs:
\begin{enumerate}
    \item A set of semantic topics\footnote{We use the term ``semantic topics'' to emphasize that these groupings are derived from item-text similarity and topic representation, not from respondent covariance or latent-variable modeling. They may be compared with theoretical or psychometric factors in subsequent evaluation, but not be interpreted as psychometric latent factors themselves.}:
    \begin{equation}
        \mathcal{T} = \{T_1,T_2,\ldots,T_K\}
    \end{equation}
    where \textit{K} is the number of topics that is automatically inferred from item semantics rather than prespecified.
    \item For each topic $T_k$, a ranked subset of representative items selected based on semantic membership stability.
    \item A low-dimensional semantic embedding space that supports visualization and inspection of item–topic relationships.
\end{enumerate}

These outputs jointly constitute a semantic pre-structure of the scale, which can subsequently guide traditional psychometric validation and refinement.

\subsubsection{Basic Workflow}
Given the formal definition above, the framework follows five sequential steps as illustrated in Figure \ref{fig:intro-framework}: encoding-dimensionality reduction, clustering, text preprocessing, topic modeling, and representative item selection.

\paragraph{Step 1-1: Encoding} Each scale item (i.e., text statement) is encoded into a dense vector representation using a pre-trained language model.
This embedding process projects the items into a high-dimensional semantic space, where geometrical proximity reflects conceptual similarity between items. \footnote{When the scale contains reverse-keyed items, they need to be rephrased into positive-keyed statements before encoding to reduce potential artifacts from negation and wording polarity. The rationale, prompt, and examples are provided in Appendices A and B.}

\paragraph{Step 1-2: Dimension reduction}
To extract the most essential information and also mitigate the curse of dimensionality, dimensionality-reduction techniques such as UMAP \parencite{mcinnes2018umap} are applied.
This step preserves both global and local semantic structure while yielding a compact space suitable for clustering and visualization.

\paragraph{Step 2: Clustering}
Items are grouped into clusters based on their semantic proximity within the embedding space.
Each cluster represents a set of conceptually similar items that may reflect a candidate content domain, subtheme, or factor-like semantic grouping.

\paragraph{Step 3: Tokenization and stop-word removal}
In parallel with embedding, item texts are tokenized into separate words. Common function words (e.g., ``a'', ``about'', ``I'', numeric tokens) are removed to enhance the interpretability of subsequent topic extraction.
This textual preprocessing enables the identification of representative keywords for each emerging topic.

\paragraph{Step 4: Topic modeling}
In parallel with the clustering, a topic description is generated for each cluster by extracting high-weight keywords from the text analysis result in step 3.
These topic labels provide interpretive summaries of semantic domains and are utilized to merge adjacent clusters to when their linguistic content substantially overlaps.

\paragraph{Step 5: Representative-item selection}
Within each finalized cluster, items are ranked by their membership strength. That is, the probability of belonging to that semantic topic.
Items with the highest membership scores are retained as representative questions, finally outputting a concise yet semantically balanced short form.

\subsubsection{Parameter Settings}
While semantic representation learning and clustering algorithms can be sensitive to model choice and parameterization, the present framework adopts empirically grounded default settings designed for short- to medium-length psychological questionnaires, where the suggestions can be found in Table \ref{tab:method-para}.

\paragraph{Encoding model}
Because all subsequent steps depend on item-level semantic distances, the sentence encoder should provide reliable representations for short texts, clustering, and multilingual semantic comparison. In the present implementation, we adopted Qwen3-Embedding-4B \parencite{zhang2025qwen3}, an open-source large language model-based embedding encoder\footnote{The proposed framework is not tied to Qwen3-Embedding-4B. Different encoders may yield somewhat different semantic neighborhoods because embedding spaces reflect model architecture, training corpora, language coverage, and cultural-linguistic exposure. We therefore regard Qwen3-Embedding-4B as a strong and reproducible default rather than as the only valid encoder. Future applications may use and compare alternative encoders when domain-specific, language-specific, or culturally sensitive item content is involved.}. This model was selected because (1) the Qwen3-Embedding series \parencite{qwen3_embedding_4b} has demonstrated strong performance over traditional sentence-transformer models (e.g., all-mpnet-base-v2, all-MiniLM-L6-v2 \parencite{sbert2026model}) on broad text-embedding benchmarks \parencite{huggingface2026mteb}, including retrieval, classification, clustering, and multilingual semantic representation tasks. These capabilities are well aligned with the present setting, where questionnaire items must be represented as short semantic units and grouped according to conceptual similarity, (2) it supports multilingual input, can be loaded through commonly used libraries such as sentence-transformers \parencite{reimers2019sentence}, and (3) the 4B size offers a practical balance between representational capacity and local deployability on consumer-level GPUs compared with larger variants.

\paragraph{Dimensionality reduction}
UMAP (uniform manifold approximation and projection) \parencite{mcinnes2018umap} was selected over linear methods (e.g., PCA) because of its ability to preserve both local and global neighborhood structure in high-dimensional semantic spaces. For questionnaires with limited item counts, reducing embeddings to approximately 5-15 dimensions provided a stable trade-off between separability and noise reduction.

\paragraph{Clustering}
Density-based clustering via HDBSCAN (hierarchical density-based spatial clustering of applications with noise) \parencite{mcinnes2017hdbscan} was used to avoid prespecifying the number of latent dimensions. In practice, parameters such as \textit{min\_cluster\_size = 3-5} and \textit{min\_samples = 1-3} for HDBSCAN yielded stable and interpretable topic structures across short to medium-size scales. These values reflect the assumption that each latent construct should be represented by at least a minimal number of semantically coherent items, while allowing peripheral or ambiguous items to remain unassigned.

\paragraph{Topic modeling}
To prevent over-fragmentation, clusters with highly overlapping keyword representations were automatically merged based on cosine similarity of their \textit{c-TF-IDF} topic representations, using a similarity threshold of 0.9. Importantly, this step prioritizes conceptual coherence over maximal partitioning.

\paragraph{Representative-item selection}
Items were ranked according to their membership probability within each topic, which considers both the semantic stability and geometric proximity. Here, empirically, an item with a probability over 0.85 (the higher the better) can be regarded as a representative of the corresponding topic.

\begin{table}[htbp]
  \centering
  \caption{Default and Recommended Parameter Settings}
    \resizebox{0.95\linewidth}{!}{
    \begin{tabular}{m{8em}<{\centering}m{23em}<{\centering}m{8em}<{\centering}}
    \toprule
    \textbf{Parameter} & \textbf{Meaning} & \textbf{Typical value used in paper} \\
    \midrule
    \textit{\textbf{UMAP}} &  &  \\
    \textit{n\_neighbors} &   Controls the balance between preserving local versus global structure by setting the number of neighboring points considered.    & 3 \\
    \textit{n\_components} &    Specifies the dimensionality of the reduced embedding space.   & 5 \\
    \textit{min\_dist} &   Determines how tightly points are packed together in the low-dimensional embedding.    & 0.0 \\
    \textit{\textbf{HDBSCAN}} &       &  \\
    \textit{min\_cluster\_size} &    Defines the minimum number of items required to form a cluster.   & 2 \\
    \textit{min\_samples} &    Controls how conservative the clustering is by influencing the identification of core versus boundary points.   & 1 \\
    \textit{\textbf{BERTopic}} &       &  \\
    \textit{nr\_topics} &    Specifies the target number of topics, either predefined based on theoretical structure or automatically determined.   & Actual factor number/+1 of scale if known, or ``auto'' \\
    \textit{top\_n\_words} &   Determines the number of top-ranked words used to represent and interpret each topic.    & 2 \\
    \bottomrule
    \end{tabular}}%
  \label{tab:method-para}%
\end{table}%

\subsection{Evaluation of the Proposed Framework}
\subsubsection{Scales}
This study selects three well-established psychological scales for evaluating the proposed method: the Depression Anxiety Stress Scales (DASS), the International Personality Item Pool (IPIP), and the Chinese version of the EPOCH Measure of Adolescent Well-Being (EPOCH-CN), the details are given in Table \ref{tab:eva-data}. The reasons for choosing them are as follows:
\begin{enumerate}
    \item All three scales are widely used and well-validated, ensuring that their theoretical structures and psychometric properties are well understood. They also have large and publicly available empirical datasets that enable robust evaluation and reproducible analyses afterwards.
    \item They span distinct construct domains, including negative affect, personality traits, and positive well-being, allowing the framework to be examined across heterogeneous psychological content.
    \item Semantically, these scales differ systematically in item valence and phrasing style: DASS consists predominantly of negatively worded items, EPOCH uses exclusively positively framed items, and IPIP contains a mixture of positively and negatively keyed statements. This variation provides a stringent test of whether semantic structure discovery is robust to surface-level linguistic polarity.
    \item Finally, we deliberately include the Chinese version of the EPOCH measure, together with data collected from Chinese participants, to provide an initial test of the effectiveness, the cross-language and -cultural applicability of the proposed framework.
\end{enumerate}

\begin{table}[htbp]
  \centering
  \caption{Information of Scales and Response Data for Evaluation}
    \resizebox{0.95\linewidth}{!}{
    \begin{tabular}{m{5em}<{\centering}m{7em}<{\centering}m{5em}<{\centering}m{8em}<{\centering}m{10em}<{\centering}m{5em}<{\centering}}
    \toprule
    \textbf{Scale name} & \textbf{Scale origin} & \textbf{Number of items} & \textbf{Number and names of the factors} & \textbf{Data origin} & \textbf{Number of data records} \\
    \midrule
    \textit{DASS} &   \parencite{lovibond1995manual}    & 42    & 3, Depression, Anxiety, Stress    &   \parencite{openpsychometrics_dass_2019}    & 39,775 \\
    \textit{IPIP} &   \parencite{goldberg1999broad}    & 50    & 5,  Extraversion, Agreeableness, Conscientiousness, Emotional Stability, Intellect    &   \parencite{openpsychometrics_ipip_ffm_2018}    & 1,015,341* \\
    \textit{EPOCH (Chinese version)} &   \parencite{kern2016epoch}(English), \parencite{kern2019chepoch}(Chinese)   & 20    & 5, Engagement, Perseverance, Optimism, Connectedness, Happiness.     &    \parencite{zeng2019network, zeng2019epoch_data}  & 17,854 \\
    \bottomrule
    \multicolumn{6}{l}{\footnotesize *: We filtered records with missing item(s), and randomly allocated 20,000 data for evaluation. All reversed items were}\\
    \multicolumn{6}{l}{\footnotesize rewritten into their counterparts prior to analysis to prevent the encoding and topic modeling from being driven by}\\
    \multicolumn{6}{l}{\footnotesize superficial polarity cues than substantive item meaning (refer to Appendix A - IPIP for original and rephrased items,}\\ \multicolumn{6}{l}{\footnotesize Appendix B for rephrasing prompt).}

    \end{tabular}}%
  \label{tab:eva-data}%
\end{table}%

\subsubsection{Evaluation Dimensions}
The evaluation is mainly organized along two complementary dimensions: psychometric evidence and methodological evidence. Additionally, robustness analyses under parameter perturbation is also considered.

\paragraph{Psychometric Evidence}
To evaluate whether the semantically derived short forms retain essential measurement properties, we assess structural validity and internal consistency using standard psychometric criteria. Importantly, these analyses are conducted post hoc and serve solely as validation, which means no data or these results would inform or constrain the semantic item selection process.

First, structural validity is examined by confirmatory factor analysis (CFA). This comparison assesses whether semantic topic–based item selection preserves theoretically meaningful latent structure rather than collapsing content into an undifferentiated general factor. Model fit is evaluated using commonly reported indices, including the Comparative Fit Index (CFI), Tucker–Lewis Index (TLI), and the Root Mean Square Error of Approximation (RMSEA).
Second, Internal consistency is examined using Cronbach’s $\alpha$ at both the total-scale and subscale levels. In addition, corrected item-total correlations (CITC) are inspected to evaluate the contribution of individual items to their respective dimensions. These indices provide evidence of whether the retained items function coherently as indicators of the underlying constructs.

\paragraph{Methodological Evidence}
The methodological analyses address whether the semantic structures discovered from item texts are theoretically meaningful, interpretable, and capable of preserving substantive information, independent of response data.

First, alignment between topic modeling result and theoretical factors. A central methodological question is: whether the semantic clusters identified by the proposed framework align with established theoretical dimensions of psychological constructs? To examine this alignment, cluster assignments produced by the semantic topic-modeling procedure were compared with the original factor labels defined by each questionnaire’s theoretical framework. 
The metric used here is Adjusted Rand Index (ARI), which measures agreement between two partitions while correcting for chance. In addition, for descriptive purposes, each semantic cluster was mapped to its dominant theoretical factor based on majority item membership. The ARI is defined as follows:

\begin{equation}
	\text{ARI} = \frac{\sum_{ij} \binom{n_{ij}}{2} - \left[ \sum_{i} \binom{a_{i}}{2} \sum_{j} \binom{b_{j}}{2} \right] / \binom{n}{2}}{\frac{1}{2} \left[ \sum_{i} \binom{a_{i}}{2} + \sum_{j} \binom{b_{j}}{2} \right] - \left[ \sum_{i} \binom{a_{i}}{2} \sum_{j} \binom{b_{j}}{2} \right] / \binom{n}{2}}
\end{equation}

\noindent where \(n_{ij}\) is the number of elements in the intersection of clusters \(i\) and \(j\), \(a_{i}\) is the sum over row \(i\), and \(b_{j}\) is the sum over column \(j\).

Second, coverage and information fidelity. We evaluated the fidelity from another perspective, as whether the shortened scales preserved substantive information contained in the original instruments by two complementary sub-aspects.
The first sub-aspect is the pattern of inter-factor correlations obtained from the shortened scales with those derived from the full-length scales, assessing whether the relational structure among latent dimensions was preserved. Structural similarity was also quantified using the normalized Frobenius similarity between the full and short subscale correlation matrices. In the second sub-aspect, we examined correlations between total and subscale scores of the original and shortened scales. 
These analyses assess information fidelity without requiring score rescaling or transformation, and provide evidence that semantic reduction retains meaningful variance present in the original measures.

Third, semantic structure visualization. To enhance interpretability and methodological transparency, the semantic embedding space of items was visualized using two-dimensional t-SNE projections. Each item was positioned according to its embedding vector and colored based on its predefined theoretical factor. In parallel, topic modeling assignments were represented by cluster boundaries drawn in the projected space.
Two complementary boundary representations were employed. Firstly, for each topic cluster, a convex hull was constructed to capture the geometric extent of items assigned to that cluster. Given a set of projected points $\{ x_i \}_{i=1}^{n}$, the convex hull corresponds to the minimal convex polygon enclosing all cluster members, thereby reflecting the outer semantic limits of the cluster:

\begin{equation}
    \mathrm{Conv}\left( \{ x_i \}_{i=1}^{n} \right)
=
\left\{
\sum_{i=1}^{n} \lambda_i x_i
\;\middle|\;
\lambda_i \ge 0,\;
\sum_{i=1}^{n} \lambda_i = 1
\right\}
\end{equation}

\noindent Secondly, a covariance-based ellipse was overlaid to approximate the statistical dispersion of each cluster. Specifically, for a cluster with mean vector $\mu$ and covariance matrix $\Sigma$, the ellipse corresponds to the contour:

\begin{equation}
\phi = \left\{
x \in \mathbbm{R}^2
\;\middle|\;
(x - \mu)^{\top} \Sigma^{-1} (x - \mu) = c
\right\}
\end{equation}

\noindent where $c$ controls the scale (e.g., proportional to the chosen standard deviation multiplier). This elliptical boundary summarizes the central tendency and principal directions of variation within the cluster.

These visualizations allow qualitative examination of (a) the alignment between theoretical factors and data-driven topic clusters, (b) the internal coherence and separation of clusters in embedding space, and (c) whether selected items occupy central or structurally representative positions within their respective clusters.

\subsection{Parameter Setting Guidance and Stability Analysis}
To provide researchers with an intuitive sense of how the proposed framework behaves under different configurations, and how to control the degree of parameter settings in their situations, we further conducted a series of targeted parameter sensitivity analyses. The goal of these evaluations was not to identify a single optimal setting, but to illustrate the scale and direction of parameter effects and to support informed, user-guided configuration in applied settings.

We focused on two groups of parameters. First, we examined two structurally decisive parameters: the number of semantic topics and the number of items retained per topic, which directly shape the semantic partitioning of items and the composition of the simplified scale. Here, using the IPIP dataset as an illustrative example, we evaluated how varying these parameters affected the structural quality of the resulting short forms, as reflected by confirmatory factor analysis indices (CFI and TLI).

Second, despite having recommended keeping other parameters related to dimensionality reduction and clustering (e.g., \textit{n\_neighbors}, \textit{min\_cluster\_size}, and \textit{min\_samples}) at their default values in most applications (as shown in Table \ref{tab:method-para}), we also conducted an additional robustness check across all three scale-datasets. These parameters were perturbed within a reasonable range, and stability was assessed by examining the overlap of selected items across settings using the Jaccard similarity index, which is defined as:

\begin{equation}
J(S_{\mathrm{default}}, S_{\mathrm{perturbed}})
= \frac{\left| S_{\mathrm{default}} \cap S_{\mathrm{perturbed}} \right|}
{\left| S_{\mathrm{default}} \cup S_{\mathrm{perturbed}} \right|},
\end{equation}

\noindent where $S$ denotes the set of selected representative items.

Together, these analyses characterize the sensitivity and stability of the framework across key configuration choices. By distinguishing parameters that primarily affect semantic granularity from those that mainly influence selection stability, this evaluation aims to help researchers develop a practical understanding of how different settings shape the simplification process and its outcomes, making it more feasible when using the simplification algorithm/tool provided.

\subsection{Software and Computational Environment}
Python 3.12.11 \parencite{python312} was used as the basic programming framework in the questionnaire simplification procedure. The large language encoding framework and realization were built upon PyTorch version 2.7.1 (basic model framework, \parencite{paszke2019pytorch}), Transformers version 4.54.0 (language model loading, \parencite{wolf2020transformers}), and Sentence Transformers version 5.0.0 (item sentence embedding acquisition, \parencite{reimers2019sentence}). The dimension reduction tool was implemented using the umap-python library version 0.5.9 \parencite{mcinnes2018umap}, while the clustering was conducted using the hdbscan library version 0.8.40 \parencite{mcinnes2017hdbscan}. The topic-modeling and -fitting framework was achieved by bertopic version 0.17.3 \parencite{grootendorst2022bertopic}. Besides, SPSS version 26.0 \parencite{ibm_spss_2019} was employed for descriptive analyses, correlation tests, assessments of internal consistency reliability, and Mplus version 8.3 \parencite{mplus_2017} was used to conduct CFA. All implementation details are reported to facilitate replication and to support the use of the proposed framework as a transparent and reusable methodological tool.

To support reproducibility, all stochastic components in the analytical pipeline (e.g., dimensionality reduction and visualization) were controlled using a fixed random seed \textit{42}. Across repeated runs with identical settings, the topic modeling and item selection results remained stable, indicating that the proposed framework yields consistent solutions under fixed configurations.

\newpage

\section{Results}
In this section, we report results in the following order: first, psychometric properties of the simplified scales; second, methodological evidence regarding semantic structure discovery; and finally, robustness analyses under parameter perturbation.

\subsection{Psychometric Results of the Simplified Scales}
\subsubsection{Depression Anxiety Stress Scales (DASS)}
The proposed semantic framework was first evaluated on the DASS scale. Using only item texts as input, the proposed framework analyzed and finally selected a shortened version comprising twelve items out of forty-two items, with four items representing each latent dimension. The detailed item selection result and representative keywords are shown in Table \ref{tab:result-dass}. The detailed scale content can be found in appendix A.

\begin{table}[htbp]
  \centering
  \caption{Scale Simplification Result of DASS}
    \begin{tabular}{cccc}
    \toprule
    \multicolumn{1}{l}{Topic No.} & Corresponding Factor & Selected Items & Topic Keywords (Top 3) \\
    \midrule
    0     & Anxiety & Q4, Q7, Q20, Q25 & absence, physical, heart \\
    1     & Depression & Q3, Q5, Q10, Q13 & feel, couldn, life \\
    2     & Stress & Q1, Q6, Q8, Q11 & get, upset, hard \\
    \bottomrule
    \end{tabular}%
  \label{tab:result-dass}%
\end{table}%

\begin{figure}[ht]
\centering
    \caption{CFA Graph of DASS (three-factor model)}
    \includegraphics[width=0.5\linewidth]{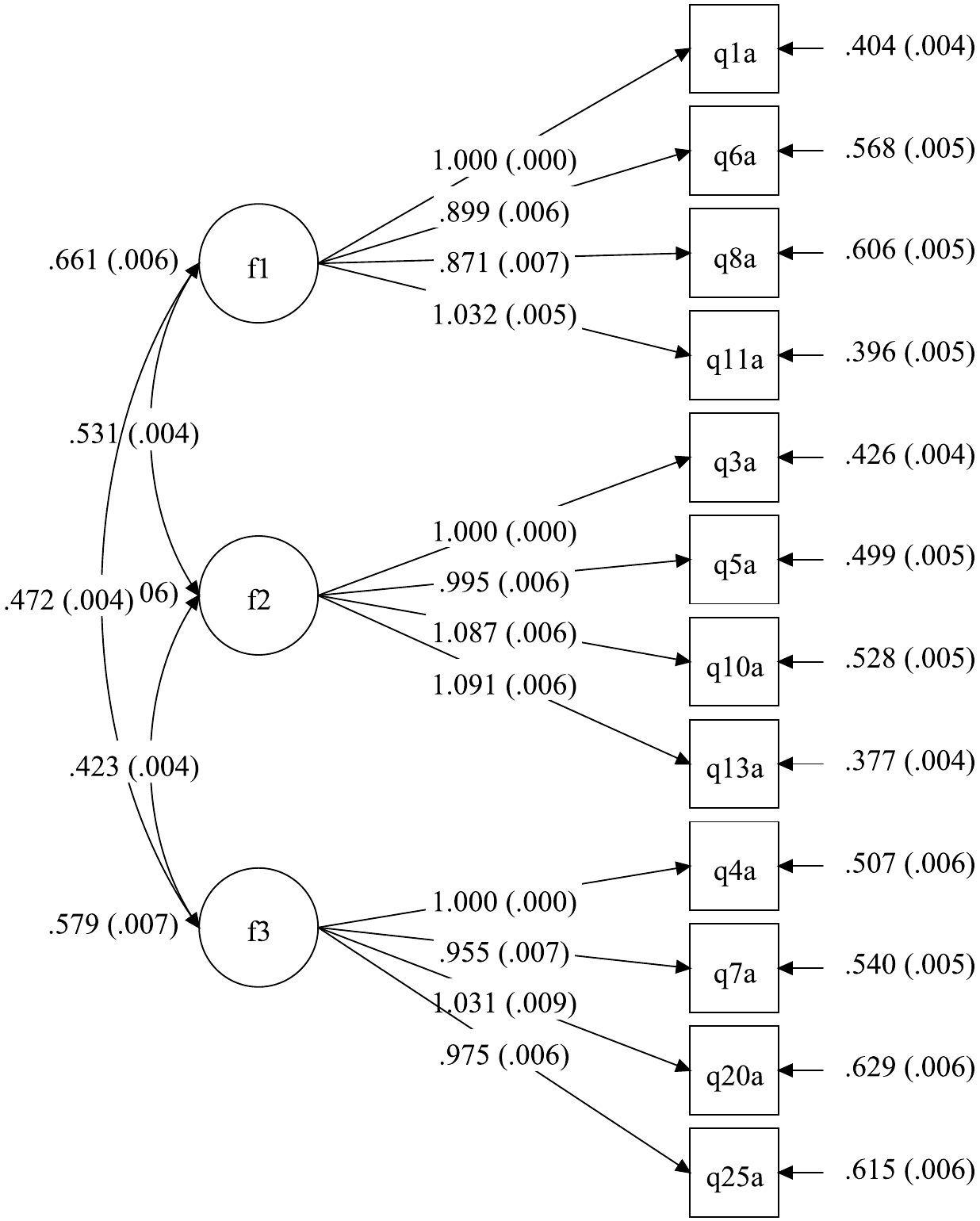}
    \label{fig:cfa-dass-3}
\end{figure}

\begin{figure}[ht]
\centering
    \caption{CFA Graph of DASS (one-factor model)}
    \includegraphics[width=0.5\linewidth]{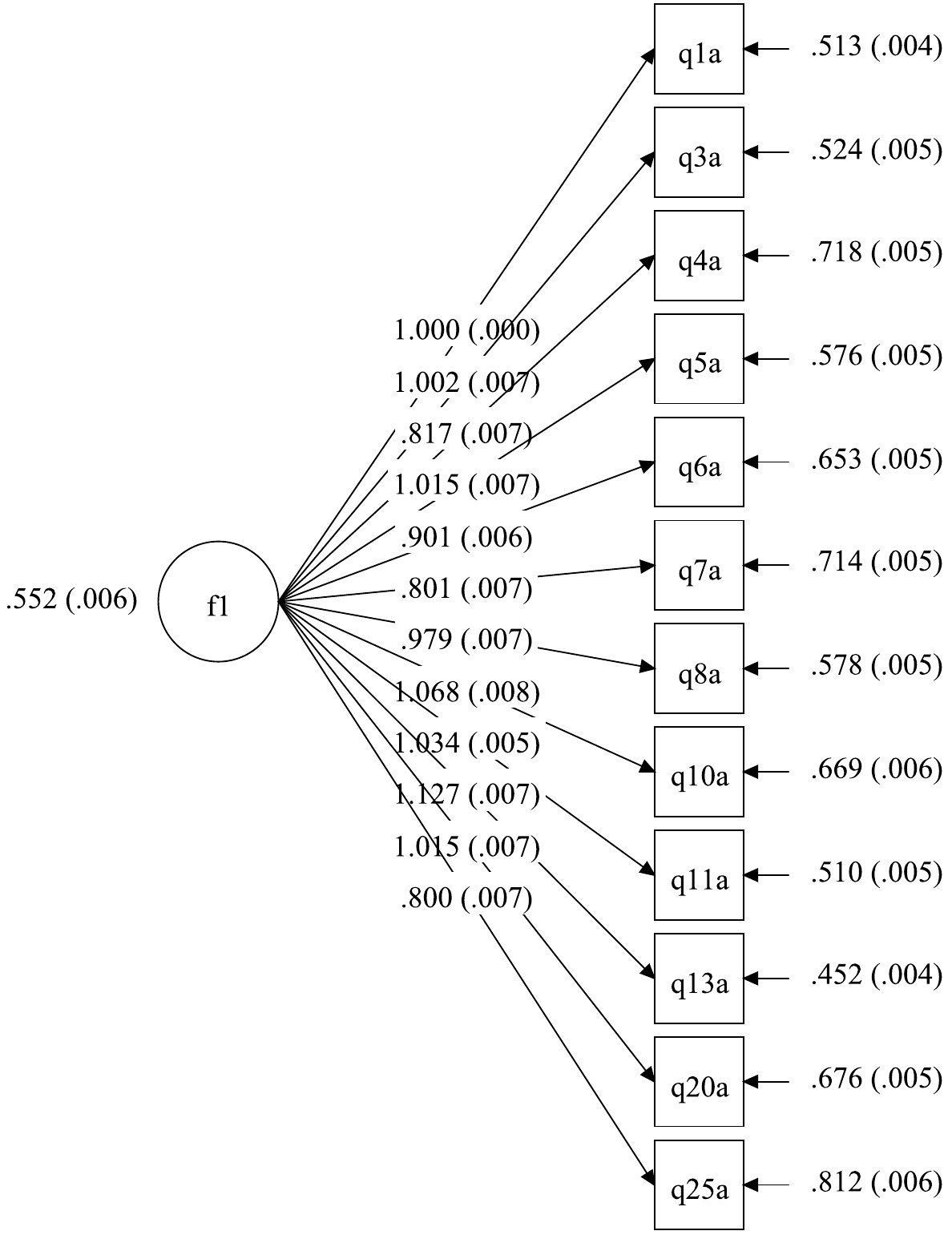}
    \label{fig:cfa-dass-1}
\end{figure}

\paragraph{Structural Validity} 
CFA was conducted to examine whether the original three-factor structure of the DASS was preserved in the shortened scale, results are shown in Tables \ref{tab:re-fact-ana-dass}, \ref{tab:re-fact-load-dass} and Figures \ref{fig:cfa-dass-3}, \ref{fig:cfa-dass-1}. The three-factor model demonstrated an acceptable to good fit to the data, $\chi^2$(51) = 8354.955, RMSEA = 0.064, CFI = 0.956, TLI = 0.943, SRMR = 0.041. In contrast, a competing single-factor model showed substantially poorer fit, $\chi^2$(54) = 26483.266, RMSEA = 0.111, CFI = 0.859, TLI = 0.828, SRMR = 0.061. These results indicate that the multidimensional structure of the DASS was retained in the semantically derived short form.
All standardized factor loadings in the three-factor model were high and statistically significant (ps < .001), ranging from 0.673 to 0.821. Items selected to represent each dimension loaded strongly on their intended factors, suggesting that the semantic selection procedure preserved clear construct differentiation.

\begin{table}[htbp]
  \centering
  \caption{Confirmatory Factor Analysis Fit Indices for the Shortened DASS}
    \begin{tabular}{lccccccc}
    \toprule
    Model & \multicolumn{1}{l}{$\chi^2$} & \multicolumn{1}{l}{df} & \multicolumn{1}{l}{$\chi^2$/df} & \multicolumn{1}{l}{RMSEA} & \multicolumn{1}{l}{CFI} & \multicolumn{1}{l}{TLI} & \multicolumn{1}{l}{SRMR} \\
    \midrule
    Three-factor & 8354.955 & 51    & 163.823 & 0.064 & 0.956 & 0.943 & 0.041 \\
    One-factor & 26483.266 & 54    & 490.431 & 0.111 & 0.859 & 0.828 & 0.061 \\
    \bottomrule
    \end{tabular}%
  \label{tab:re-fact-ana-dass}%
\end{table}%

\begin{table}[htbp]
  \centering
  \caption{Factor Loadings of the Shortened DASS (\textit{n} = 39,775)}
    \begin{tabular}{lrlrlr}
    \toprule
    \multicolumn{2}{c}{Stress} & \multicolumn{2}{c}{Depression} & \multicolumn{2}{c}{Anxiety}   \\
    \midrule
    Item  & \multicolumn{1}{l}{Factor Loading} & Item  & \multicolumn{1}{l}{Factor Loading} & Item  & \multicolumn{1}{l}{Factor Loading} \\
    \midrule
    Q1    & 0.788 & Q3    & 0.778 & Q4    & 0.730 \\
    Q6    & 0.696 & Q5    & 0.751 & Q7    & 0.703 \\
    Q8    & 0.673 & Q10   & 0.770 & Q20   & 0.703 \\
    Q11   & 0.800 & Q13   & 0.821 & Q25   & 0.687 \\
    \bottomrule
    \end{tabular}%
  \label{tab:re-fact-load-dass}%
\end{table}%

\paragraph{Item Quality and Internal Consistency}
Item-level analyses further supported the quality of the shortened DASS. As shown in Table \ref{tab:re-real-dass}, the CITC values ranged from 0.553 to 0.719, with no items falling below commonly accepted thresholds for adequate discrimination, indicating strong associations between individual items and their respective subscale scores.
Internal consistency was high for the total scale (Cronbach’s $\alpha$ = 0.899). Subscale reliabilities were also satisfactory, with Cronbach’s $\alpha$ values of 0.824 for the Depression, 0.862 for the Anxiety, and 0.797 for the Stress subscale. Together, these results indicate that the shortened DASS maintains strong internal coherence at both the total-scale and subscale levels.

\begin{table}[htbp]
  \centering
  \caption{Item Quality and Internal Consistency of the Shortened DASS}
    \begin{tabular}{ccrccr}
    \toprule
    \multicolumn{2}{c}{Stress} & \multicolumn{2}{c}{Depression} & \multicolumn{2}{c}{Anxiety}  \\
    \midrule
    \multicolumn{1}{l}{Item} & \multicolumn{1}{l}{CITC} & \multicolumn{1}{l}{Item} & \multicolumn{1}{l}{CITC} & \multicolumn{1}{l}{Item} & \multicolumn{1}{l}{CITC} \\
    \multicolumn{1}{l}{Q1} & \multicolumn{1}{r}{0.696} & \multicolumn{1}{l}{Q3} & \multicolumn{1}{r}{0.714} & \multicolumn{1}{l}{Q4} & 0.653 \\
    \multicolumn{1}{l}{Q6} & \multicolumn{1}{r}{0.639} & \multicolumn{1}{l}{Q5} & \multicolumn{1}{r}{0.684} & \multicolumn{1}{l}{Q7} & 0.619 \\
    \multicolumn{1}{l}{Q8} & \multicolumn{1}{r}{0.556} & \multicolumn{1}{l}{Q10} & \multicolumn{1}{r}{0.719} & \multicolumn{1}{l}{Q20} & 0.553 \\
    \multicolumn{1}{l}{Q11} & \multicolumn{1}{r}{0.706} & \multicolumn{1}{l}{Q13} & \multicolumn{1}{r}{0.719} & \multicolumn{1}{l}{Q25} & 0.614 \\
    \midrule
    \multicolumn{1}{l}{Cronbach’s $\alpha$} & \multicolumn{1}{r}{0.824} & \multicolumn{1}{l}{Cronbach’s $\alpha$} & \multicolumn{1}{r}{0.862} & \multicolumn{1}{l}{Cronbach’s $\alpha$} & 0.797 \\
    \multicolumn{2}{c}{Cronbach’s $\alpha$ (overall)} &       & \multicolumn{2}{c}{0.899} &  \\
    \bottomrule
    \end{tabular}%
  \label{tab:re-real-dass}%
\end{table}%

\subsubsection{International Personality Item Pool (IPIP)}
The semantic framework was further evaluated on the IPIP scale. 
The detailed item selection result and representative keywords are shown in Table \ref{tab:result-ipip}. The complete scale content is presented in appendix A.

\begin{table}[htbp]
  \centering
  \caption{Scale Simplification Result of IPIP}
    \resizebox{0.95\linewidth}{!}{
    \begin{tabular}{m{5em}<{\centering}m{8em}<{\centering}m{10em}<{\centering}m{10em}<{\centering}}
    \toprule
    \multicolumn{1}{l}{Topic No.} & Corresponding Factor & Selected Items & Topic Keywords (Top 3) \\
    \midrule
    0     & EXT (Extraversion) & Q1, Q2, Q3, Q5 (original item ids: EXT \{1,2,3,5\}) & people, talk, attention \\
    1     & EST (Emotional Stability, or Neuroticism) & Q12, Q13, Q14, Q16 (EST \{2,3,4,6\}) & stay, calm, relax \\
    2     & CSN (Conscientiousness) & Q31, Q32, Q33, Q34 (CSN \{1,2,3,4\}) & order, thing, keep \\
    3     & AGR (Agreeableness) & Q24, Q25, Q26, Q28 (AGR \{4,5,6,8\}) & feel, people, other \\
    4     & OPN (Openness, or Intellect) &  Q42, Q43, Q45, Q46 (OPN \{2,3,5,6\}) & idea, understand, imagination \\
    \bottomrule
    \end{tabular}}%
  \label{tab:result-ipip}%
\end{table}%

\begin{figure}[ht]
\centering
    \caption{CFA Graph of IPIP (five-factor model)}
    \includegraphics[width=0.5\linewidth]{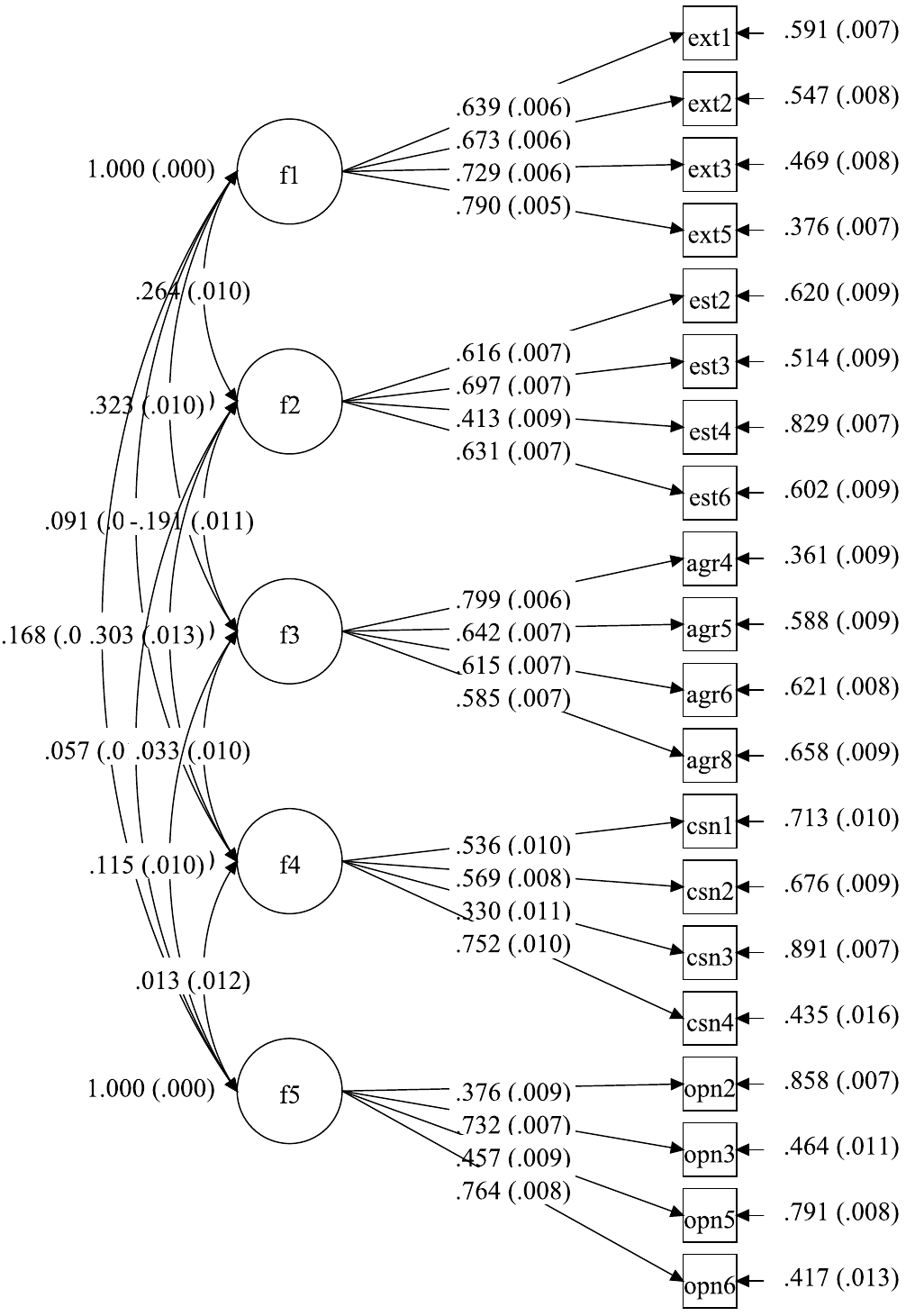}
    \label{fig:cfa-ipip-5}
\end{figure}

\begin{figure}[ht]
\centering
    \caption{CFA Graph of IPIP (one-factor model)}
    \includegraphics[width=0.5\linewidth]{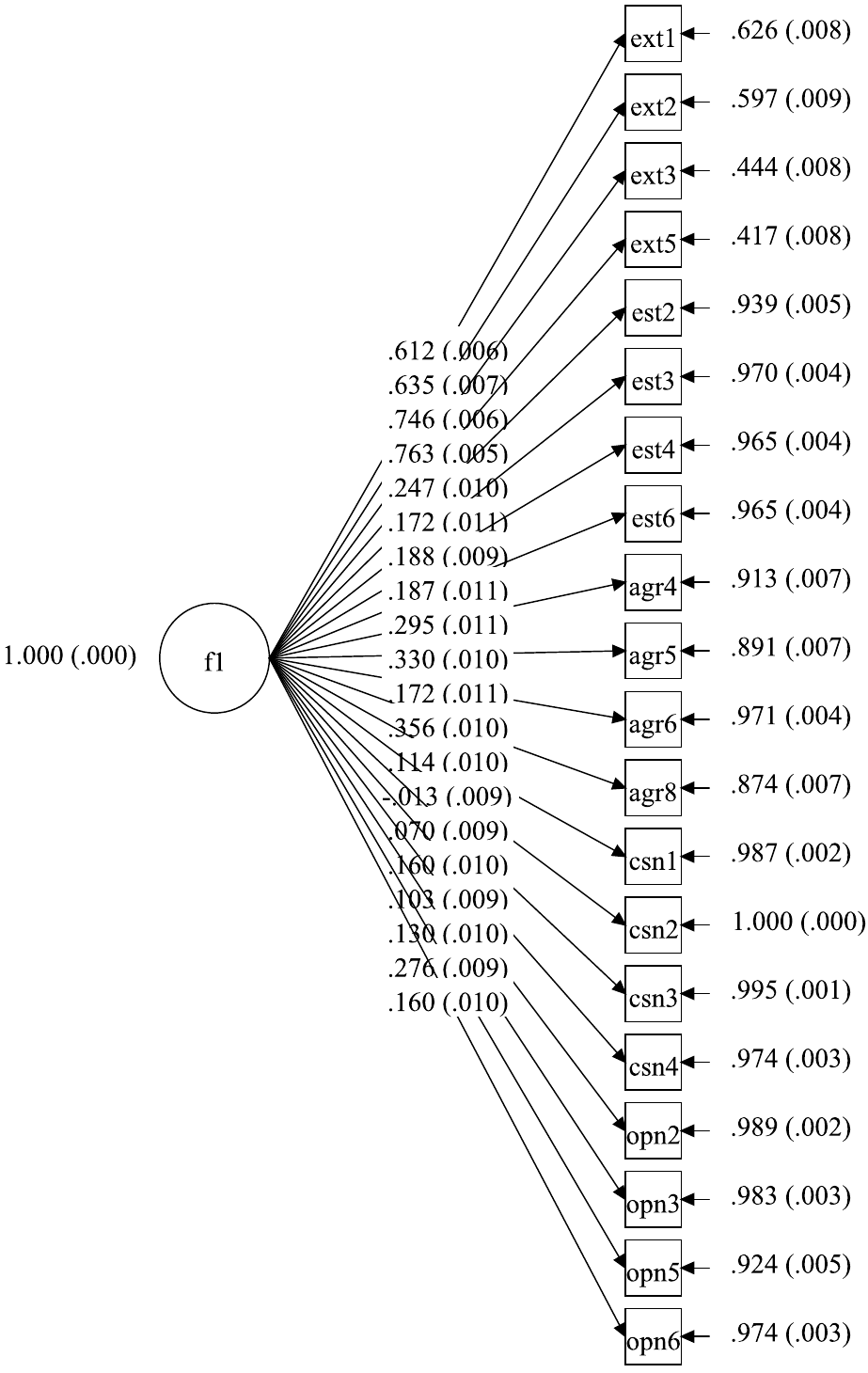}
    \label{fig:cfa-ipip-1}
\end{figure}

\paragraph{Structural Validity}
In terms of the CFA results (Tables \ref{tab:re-fact-ana-ipip},\ref{tab:re-fact-load-ipip} and Figures \ref{fig:cfa-ipip-5},\ref{fig:cfa-ipip-1}), the hypothesized five-factor model demonstrated acceptable absolute fit but weaker incremental fit indices, $\chi^2$(160) = 11,258.01, RMSEA = 0.059, CFI = 0.860, TLI = 0.833, SRMR = 0.057. In contrast, a competing single-factor model exhibited substantially poorer fit, $\chi^2$(173) = 53,927.25, RMSEA = 0.125, CFI = 0.320, TLI = 0.253, SRMR = 0.129. These results indicate that the multidimensional structure of the IPIP was largely retained in the shortened version, although CFI and TLI values should be interpreted as marginal under stricter fit criteria.
Standardized factor loadings in the five-factor model were statistically significant (ps < .001) and predominantly moderate to high, with most values exceeding 0.50. Two items exhibited relatively lower loadings (CSN3 0.330 and OPN2 0.376), reflecting the increased heterogeneity and reduced redundancy inherent in brief personality measures.\footnote{As a descriptive benchmark, the same five-factor CFA applied to the original 50-item IPIP in the present sample also showed weak incremental fit (CFI = 0.748, TLI = 0.735, SRMR = 0.060). In addition, some weaker retained items were already modest in the full form; for example, CSN3 had a loading of 0.385 in the original IPIP, close to its loading in the short form (0.330), whereas OPN2 decreased from 0.565 in the full form to 0.376 in the short form. These comparisons suggest that the weaker indices partly reflect characteristics of the original item pool and the present sample.}

\begin{table}[htbp]
  \centering
  \caption{Confirmatory Factor Analysis Fit Indices for the Shortened IPIP}
    \begin{tabular}{lccccccc}
    \toprule
    Model & \multicolumn{1}{l}{$\chi^2$} & \multicolumn{1}{l}{df} & \multicolumn{1}{l}{$\chi^2$/df} & \multicolumn{1}{l}{RMSEA} & \multicolumn{1}{l}{CFI} & \multicolumn{1}{l}{TLI} & \multicolumn{1}{l}{SRMR} \\
    \midrule
    Five-factor & 11258.01 & 160   & 70.363 & 0.059 & 0.860  & 0.833 & 0.057 \\
    One-factor & 53927.25 & 173   & 311.718 & 0.125 & 0.320  & 0.253 & 0.129 \\
    \bottomrule
    \end{tabular}%
  \label{tab:re-fact-ana-ipip}%
\end{table}%

\begin{sidewaystable}[ph!]
  \centering
  \caption{Factor Loadings of the Shortened IPIP (\textit{n} = 20,000)}
    \begin{tabular}{lrlrlrlrlr}
    \toprule
    \multicolumn{2}{c}{Extraversion} & \multicolumn{2}{c}{Emotional Stability} & \multicolumn{2}{c}{Agreeableness} & \multicolumn{2}{c}{Conscientiousness} & \multicolumn{2}{c}{Intellect} \\
    \midrule
    Item  & \multicolumn{1}{l}{Factor loading} & Item  & \multicolumn{1}{l}{Factor loading} & Item  & \multicolumn{1}{l}{Factor loading} & Item  & \multicolumn{1}{l}{Factor loading} & Item  & \multicolumn{1}{l}{Factor loading} \\
    \midrule
    EXT1  & 0.639 & EST2  & 0.616 & AGR4  & 0.799 & CSN1  & 0.536 & OPN2  & 0.376 \\
    EXT2  & 0.673 & EST3  & 0.697 & AGR5  & 0.642 & CSN2  & 0.569 & OPN3  & 0.732 \\
    EXT3  & 0.729 & EST4  & 0.413 & AGR6  & 0.615 & CSN3  & 0.330 & OPN5  & 0.457 \\
    EXT5  & 0.790 & EST6  & 0.631 & AGR8  & 0.585 & CSN4  & 0.752 & OPN6  & 0.764 \\
    \bottomrule
    \end{tabular}%
  \label{tab:re-fact-load-ipip}%
\end{sidewaystable}%

\paragraph{Item Quality and Internal Consistency}
As shown in Table \ref{tab:re-real-ipip}, the CITC values for all retained items of the IPIP exceeded 0.30, with the majority above 0.40. Three items (EST4, CSN3 and OPN2) fell within the 0.30–0.40 range, suggesting comparatively weaker item discrimination.
Internal consistency estimates varied across personality dimensions. Cronbach’s $\alpha$ coefficients ranged from 0.637 to 0.799 across the five subscales, while the overall scale demonstrated an internal consistency of $\alpha = 0.705$. These values are consistent with prior research on short-form personality inventories \parencite{goldberg1999broad} and reflect the trade-off between scale brevity and reliability.

\begin{sidewaystable}[ph!]
  \centering
  \caption{Item Quality and Internal Consistency of the Shortened IPIP}
    \begin{tabular}{ccrrrccrrr}
    \toprule
    \multicolumn{2}{c}{Extraversion} & \multicolumn{2}{c}{Emotional Stability} & \multicolumn{2}{c}{Agreeableness} & \multicolumn{2}{c}{Conscientiousness} & \multicolumn{2}{c}{Intellect} \\
    \midrule
    \multicolumn{1}{l}{Item} & \multicolumn{1}{l}{CITC} & \multicolumn{1}{l}{Item} & \multicolumn{1}{l}{CITC} & \multicolumn{1}{l}{Item} & \multicolumn{1}{l}{CITC} & \multicolumn{1}{l}{Item} & \multicolumn{1}{l}{CITC} & \multicolumn{1}{l}{Item} & \multicolumn{1}{l}{CITC} \\
    \multicolumn{1}{l}{EXT1} & \multicolumn{1}{r}{0.578} & \multicolumn{1}{l}{EST2} & 0.476 & \multicolumn{1}{l}{AGR4} & \multicolumn{1}{r}{0.649} & \multicolumn{1}{l}{CSN1} & 0.452 & \multicolumn{1}{l}{OPN2} & 0.367 \\
    \multicolumn{1}{l}{EXT2} & \multicolumn{1}{r}{0.593} & \multicolumn{1}{l}{EST3} & 0.508 & \multicolumn{1}{l}{AGR5} & \multicolumn{1}{r}{0.524} & \multicolumn{1}{l}{CSN2} & 0.424 & \multicolumn{1}{l}{OPN3} & 0.534 \\
    \multicolumn{1}{l}{EXT3} & \multicolumn{1}{r}{0.612} & \multicolumn{1}{l}{EST4} & 0.336 & \multicolumn{1}{l}{AGR6} & \multicolumn{1}{r}{0.498} & \multicolumn{1}{l}{CSN3} & 0.318 & \multicolumn{1}{l}{OPN5} & 0.425 \\
    \multicolumn{1}{l}{EXT5} & \multicolumn{1}{r}{0.665} & \multicolumn{1}{l}{EST6} & 0.472 & \multicolumn{1}{l}{AGR8} & \multicolumn{1}{r}{0.513} & \multicolumn{1}{l}{CSN4} & 0.490 & \multicolumn{1}{l}{OPN6} & 0.563 \\
    \midrule
    \multicolumn{1}{l}{Cronbach’s $\alpha$} & \multicolumn{1}{r}{0.799} & \multicolumn{1}{l}{Cronbach’s $\alpha$} & 0.664 & \multicolumn{1}{l}{Cronbach’s $\alpha$} & \multicolumn{1}{r}{0.749} & \multicolumn{1}{l}{Cronbach’s $\alpha$} & 0.637 & \multicolumn{1}{l}{Cronbach’s $\alpha$} & 0.686 \\
    \multicolumn{2}{c}{Cronbach’s $\alpha$ (overall)} &       &       &       & \multicolumn{2}{c}{0.705} &       &       &  \\
    \bottomrule
    \end{tabular}%
  \label{tab:re-real-ipip}%
\end{sidewaystable}%

\subsubsection{The Chinese Version of EPOCH Measure of Adolescent Well-Being (EPOCH-CN)}
The proposed framework was finally evaluated on the EPOCH-CN scale. The original scale consists of twenty items, resulting in relatively limited redundancy within each dimension. Based on semantic structure discovery, a shortened ten-item version was derived, as shown in Table \ref{tab:result-epoch-cn}. The detailed scale content is presented in the appendix A.

\begin{table}[htbp]
  \centering
  \caption{Scale Simplification Result of EPOCH-CN}
    \resizebox{0.95\linewidth}{!}{
    \begin{tabular}{m{5em}<{\centering}m{8em}<{\centering}m{10em}<{\centering}m{10em}<{\centering}}
    \toprule
    \multicolumn{1}{l}{Topic No.} & Corresponding Factor & Selected Items & Topic Keywords (Top 3) \\
    \midrule
    0     & E (Engagement) & Q7, Q12 (original item ids: E2, E4) & time, forget,involve \\
    1     & O (Optimism) & Q13, Q18 (O2, O4) & believe, good, thing \\
    2     & H (Happiness) & Q4, Q20 (H1, H4) & happy, fun, life \\
    3     & P (Perseverance) & Q2, Q9 (P1, P2) & finish, until, stick \\
    4     & C (Connectedness) & Q1, Q10 (C1, C2) & care, people, share \\
    \bottomrule
    \end{tabular}}%
  \label{tab:result-epoch-cn}%
\end{table}%

\begin{figure}[ht]
\centering
    \caption{CFA Graph of EPOCH-CN (five-factor model)}
    \includegraphics[width=0.5\linewidth]{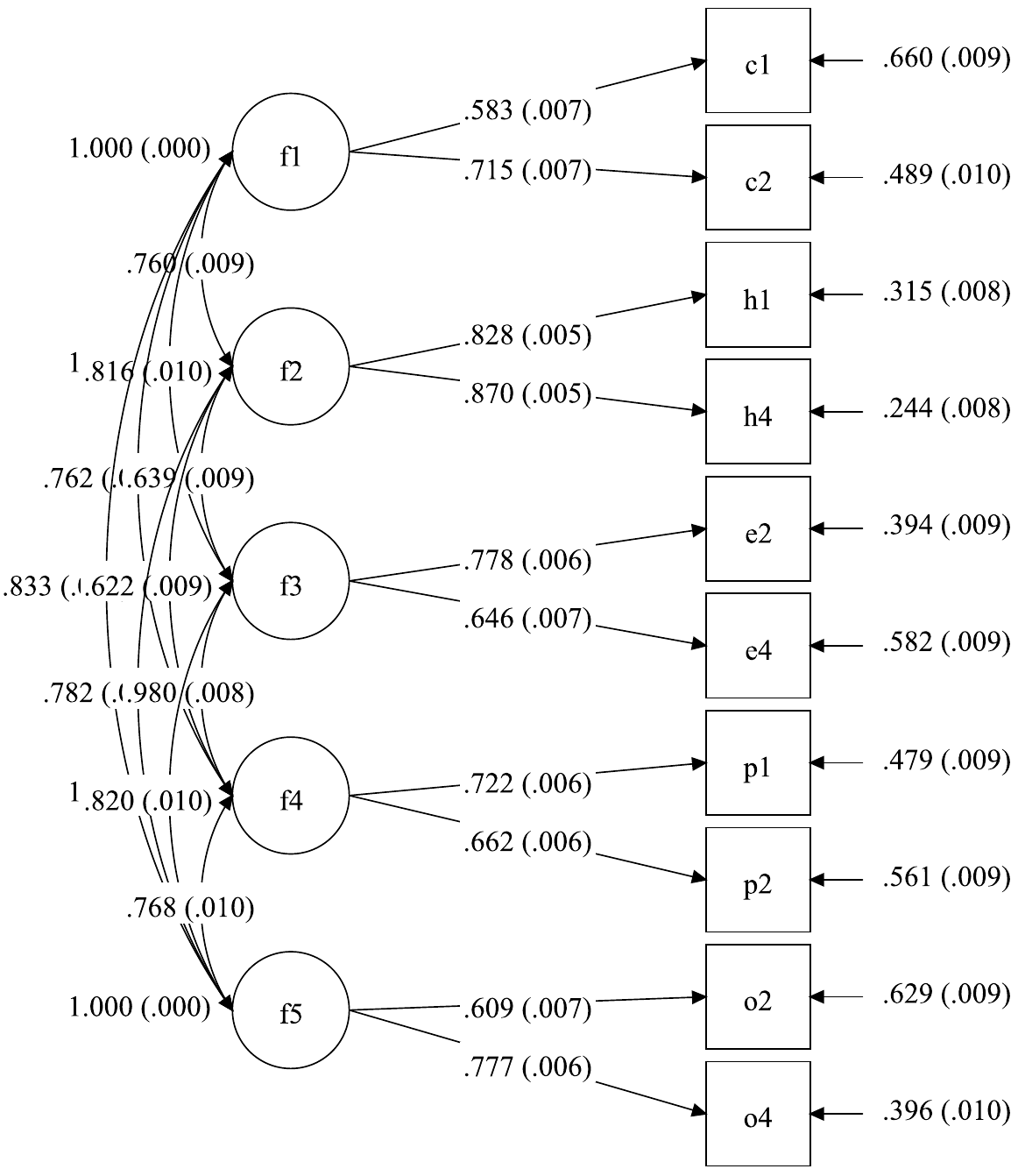}
    \label{fig:cfa-epoch-5}
\end{figure}

\begin{figure}[ht]
\centering
    \caption{CFA Graph of EPOCH-CN (one-factor model)}
    \includegraphics[width=0.5\linewidth]{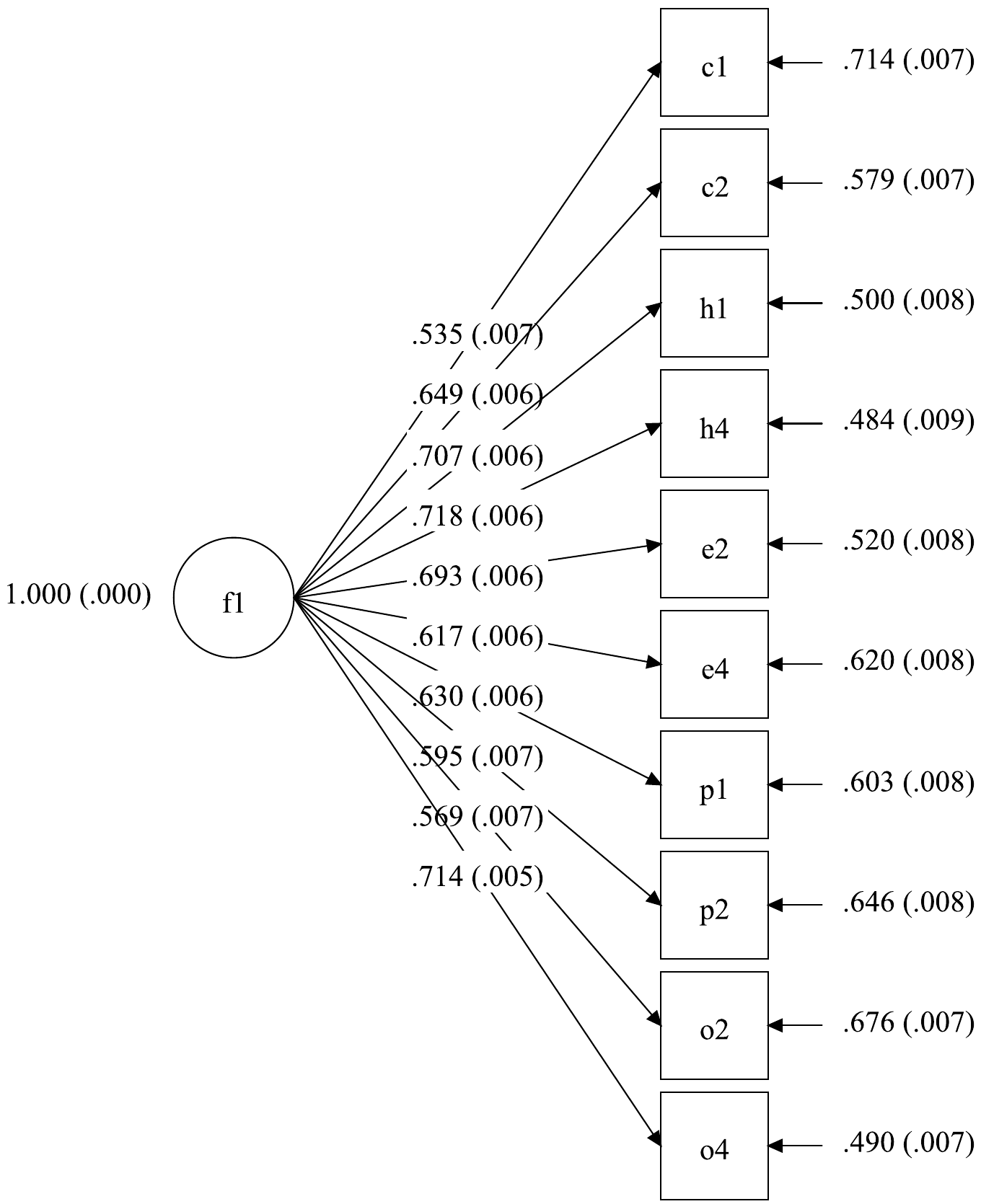}
    \label{fig:cfa-epoch-1}
\end{figure}

\paragraph{Structural Validity}
Results of CFA (Tables \ref{tab:re-fact-ana-epoch},\ref{tab:re-fact-load-epoch} and Figures \ref{fig:cfa-epoch-5},\ref{fig:cfa-epoch-1}) showed that the hypothesized five-factor model demonstrated an excellent fit to the data, with $\chi^2$(25)  = 771.534, RMSEA = 0.041, CFI = 0.983, TLI = 0.970, SRMR = 0.018, indicating strong structural validity. In contrast, a competing single-factor model showed substantially poorer fit $\chi^2$(35) = 5895.535, RMSEA = 0.097, CFI = 0.870, TLI = 0.833, SRMR = 0.050. These results indicate that the multidimensional structure of the EPOCH-CN was clearly retained despite the substantial reduction in item number.
All standardized factor loadings in the five-factor model were statistically significant (ps < .001) and exceeded 0.50. Loadings ranged from 0.583 to 0.870, with particularly strong loadings observed for items representing Happiness (0.828–0.870). The clear separation among the five latent dimensions suggests that semantic clustering was able to recover the intended well-being structure from item texts alone, even in a scale with limited item redundancy.

\begin{table}[htbp]
  \centering
  \caption{Confirmatory Factor Analysis Fit Indices for the Shortened EPOCH-CN}
    \begin{tabular}{lrrrrrrr}
    \toprule
    Model & \multicolumn{1}{l}{$\chi^2$} & \multicolumn{1}{l}{df} & \multicolumn{1}{l}{$\chi^2$/df} & \multicolumn{1}{l}{RMSEA} & \multicolumn{1}{l}{CFI} & \multicolumn{1}{l}{TLI} & \multicolumn{1}{l}{SRMR} \\
    \midrule
    Five-factor & 771.534 & 25    & 30.861 & 0.041 & 0.983 & 0.970 & 0.018 \\
    One-factor & 5895.535 & 35    & 168.444 & 0.097  & 0.870 & 0.833 & 0.050 \\
    \bottomrule
    \end{tabular}%
  \label{tab:re-fact-ana-epoch}%
\end{table}%

\begin{sidewaystable}[ph!]
  \centering
  \caption{Factor Loadings of the Shortened EPOCH-CN (\textit{n} = 17,854)}
    \begin{tabular}{lrlrlrlrlr}
    \toprule
    \multicolumn{2}{c}{Connectedness} & \multicolumn{2}{c}{Happiness} & \multicolumn{2}{c}{Engagement} & \multicolumn{2}{c}{Perseverance} & \multicolumn{2}{c}{Optimism} \\
    \midrule
    Item  & \multicolumn{1}{l}{Factor Loading} & Item  & \multicolumn{1}{l}{Factor Loading} & Item  & \multicolumn{1}{l}{Factor Loading} & Item  & \multicolumn{1}{l}{Factor Loading} & Item  & \multicolumn{1}{l}{Factor Loading} \\
    \midrule
    C1    & 0.583 & H1    & 0.828 & E2    & 0.778 & P1    & 0.722 & O2    & 0.609 \\
    C2    & 0.715 & H4    & 0.870 & E4    & 0.646 & P2    & 0.662 & O4    & 0.777 \\
    \bottomrule
    \end{tabular}%
  \label{tab:re-fact-load-epoch}%
\end{sidewaystable}%

\paragraph{Item Quality and Internal Consistency}
The CITC values for all retained items exceeded 0.40 (Table \ref{tab:re-real-epoch}). This pattern suggests generally satisfactory item quality given the extreme brevity of the scale.
The overall shortened EPOCH-CN demonstrated high internal consistency (Cronbach’s $\alpha$ = 0.875), while subscale-level reliability varied slightly across dimensions. Cronbach’s $\alpha$ coefficients were 0.837 for Happiness, 0.668 for Engagement, 0.647 for Perseverance, and 0.642 for Optimism, indicating acceptable reliability for brief subscales. In contrast, the Connectedness subscale exhibited lower internal consistency ($\alpha$ = 0.588).
This variability is consistent with well-documented psychometric constraints of two-item subscales and reflects a reliability ceiling imposed by scale length rather than deficiencies in semantic structure discovery.

\begin{sidewaystable}[ph!]
  \centering
  \caption{Item Quality and Internal Consistency of the Shortened EPOCH-CN}
    \begin{tabular}{ccrrrccrrr}
    \toprule
    \multicolumn{2}{c}{Connectedness} & \multicolumn{2}{c}{Happiness} & \multicolumn{2}{c}{Engagement} & \multicolumn{2}{c}{Perseverance} & \multicolumn{2}{c}{Optimism} \\
    \midrule
    \multicolumn{1}{l}{Item} & \multicolumn{1}{l}{CITC} & \multicolumn{1}{l}{Item} & \multicolumn{1}{l}{CITC} & \multicolumn{1}{l}{Item} & \multicolumn{1}{l}{CITC} & \multicolumn{1}{l}{Item} & \multicolumn{1}{l}{CITC} & \multicolumn{1}{l}{Item} & \multicolumn{1}{l}{CITC} \\
    \multicolumn{1}{l}{C1} & \multicolumn{1}{r}{0.417} & \multicolumn{1}{l}{H1} & 0.720 & \multicolumn{1}{l}{E2} & \multicolumn{1}{r}{0.503} & \multicolumn{1}{l}{P1} & 0.478 & \multicolumn{1}{l}{O2} & 0.473 \\
    \multicolumn{1}{l}{C2} & \multicolumn{1}{r}{0.417} & \multicolumn{1}{l}{H4} & 0.720 & \multicolumn{1}{l}{E4} & \multicolumn{1}{r}{0.503} & \multicolumn{1}{l}{P2} & 0.478 & \multicolumn{1}{l}{O4} & 0.473 \\
    \midrule
    \multicolumn{1}{l}{Cronbach’s $\alpha$} & \multicolumn{1}{r}{0.588} & \multicolumn{1}{l}{Cronbach’s $\alpha$} & 0.837 & \multicolumn{1}{l}{Cronbach’s $\alpha$} & \multicolumn{1}{r}{0.668} & \multicolumn{1}{l}{Cronbach’s $\alpha$} & 0.647 & \multicolumn{1}{l}{Cronbach’s $\alpha$} & 0.642 \\
    \multicolumn{2}{c}{Cronbach’s $\alpha$ (overall)} &       &       &       & \multicolumn{2}{c}{0.875} &       &       &  \\
    \bottomrule
    \end{tabular}%
  \label{tab:re-real-epoch}%
\end{sidewaystable}%

\subsection{Methodological Evidence}
\subsubsection{Alignment Between Semantic Clusters and Theoretical Factors}

\begin{table}[htbp]
  \centering
  \caption{Correspondence Between Semantic Clusters and DASS Factors}
    \resizebox{0.85\linewidth}{!}{
    \begin{tabular}{m{10em}<{\centering}m{9em}<{\centering}m{7em}<{\centering}m{7em}<{\centering}}
    \toprule
    \textbf{Semantic cluster} & \textbf{Dominant factor} & \multicolumn{1}{l}{\textbf{Proportion}} & \multicolumn{1}{l}{\textbf{Number of items}} \\
    \midrule
    \textit{Topic 0} & Anxiety & 0.8   & 15 \\
    \textit{Topic 1} & Depression & 1     & 14 \\
    \textit{Topic 2} & Stress & 1     & 11 \\
    \textit{Topic 3} & Anxiety & 1     & 2 \\
    \midrule
    ARI & \multicolumn{3}{c}{0.745}\\
    \bottomrule
    \end{tabular}}%
  \label{tab:re-align-dass}%
\end{table}%

\begin{table}[htbp]
  \centering
  \caption{Correspondence Between Semantic Clusters and IPIP Factors}
    \resizebox{0.9\linewidth}{!}{
    \begin{tabular}{m{10em}<{\centering}m{12em}<{\centering}m{7em}<{\centering}m{7em}<{\centering}}
    \toprule
    \textbf{Semantic cluster} & \textbf{Dominant factor} & \multicolumn{1}{l}{\textbf{Proportion}} & \multicolumn{1}{l}{\textbf{Number of items}} \\
    \midrule
    \textit{Topic 0} & Extraversion (EXT) & 0.83  & 12 \\
    \textit{Topic 1} & Emotional Stability (EST) & 1     & 10 \\
    \textit{Topic 2} & Conscientiousness (CSN) & 1     & 10 \\
    \textit{Topic 3} & Agreeableness (AGR) & 1     & 8 \\
    \textit{Topic 4} & Intellect/Openness (OPN) & 1     & 8 \\
    \textit{Topic 5} & Intellect/Openness (OPN) & 1     & 2 \\
    \midrule
    ARI & \multicolumn{3}{c}{0.855}\\
    \bottomrule
    \end{tabular}}%
  \label{tab:re-align-ipip}%
\end{table}%

\begin{table}[htbp]
  \centering
  \caption{Correspondence Between Semantic Clusters and EPOCH-CN Factors}
    \resizebox{0.85\linewidth}{!}{
    \begin{tabular}{m{10em}<{\centering}m{9em}<{\centering}m{7em}<{\centering}m{7em}<{\centering}}
    \toprule
    \textbf{Semantic cluster} & \textbf{Dominant factor} & \multicolumn{1}{l}{\textbf{Proportion}} & \multicolumn{1}{l}{\textbf{Number of items}} \\
    \midrule
    \textit{Topic 0} & Engagement (E) & 1     & 4 \\
    \textit{Topic 1} & Optimism (O) & 1     & 4 \\
    \textit{Topic 2} & Happiness (H) & 1     & 4 \\
    \textit{Topic 3} & Perseverance (P) & 1     & 4 \\
    \textit{Topic 4} & Connectedness (C) & 1     & 4 \\
    \midrule
    ARI & \multicolumn{3}{c}{1.000}\\
    \bottomrule
    \end{tabular}}%
  \label{tab:re-align-epoch}%
\end{table}%

To further examine whether the semantic structure discovered by the proposed framework corresponds to established psychometric theory, we quantified the alignment between semantic cluster assignments and the original factor structures (as ``ground truth'') of the evaluated questionnaires. Importantly, this analysis does not assess predictive accuracy, but rather evaluates the extent to which latent semantic organization mirrors theoretically defined dimensions across instruments.

For each questionnaire, cluster assignments were compared with the original factor labels using the Adjusted Rand Index (ARI), supplemented by majority-vote correspondence analyses. These metrics quantify structural agreement without requiring predefined label mappings or supervised optimization.

Across the three instruments, substantial alignment was observed between semantic clusters and theoretical factor structures, as shown in Tables \ref{tab:re-align-dass},\ref{tab:re-align-ipip} and \ref{tab:re-align-epoch}. For the DASS, which assesses negative affect across three related dimensions, semantic clustering showed strong correspondence with the Depression, Anxiety, and Stress factors (ARI = 0.745), with 92.9\% of items assigned to clusters whose dominant factor matched their theoretical classification. For the IPIP, representing a trait-based personality inventory with mixed item valence, alignment was even higher (ARI = 0.855), with 96.0\% of items exhibiting factor-consistent cluster assignments. For the EPOCH-CN, semantic clustering fully recovered the original five-factor structure (ARI = 1.000), with complete correspondence between clusters and theoretical dimensions.

Taken together, these results indicate that the proposed semantic structure discovery reliably aligns with established psychometric organization across diverse measurement domains. The observed variation in alignment strength across instruments suggests that semantic–theoretical correspondence is sensitive to construct characteristics and item composition, a pattern that is examined further in the Discussion section.

\subsubsection{Information Fidelity}
\begin{figure}[ht]
	\centering 
	\subfigure[ ]{
		\label{d-level.sub.1}
		\includegraphics[width=0.4\linewidth]{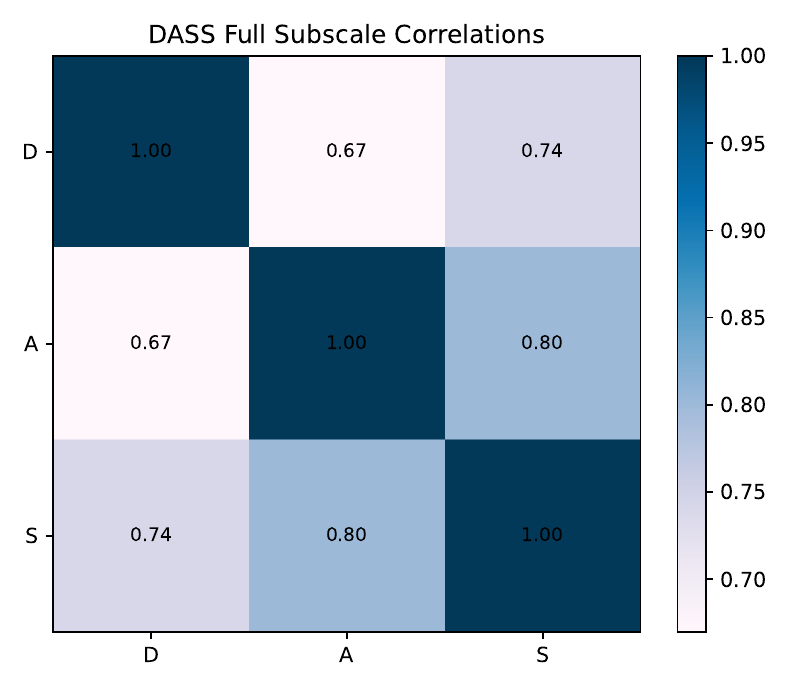}}
	\quad 
	\subfigure[ ]{
		\label{d-level.sub.2}
		\includegraphics[width=0.4\linewidth]{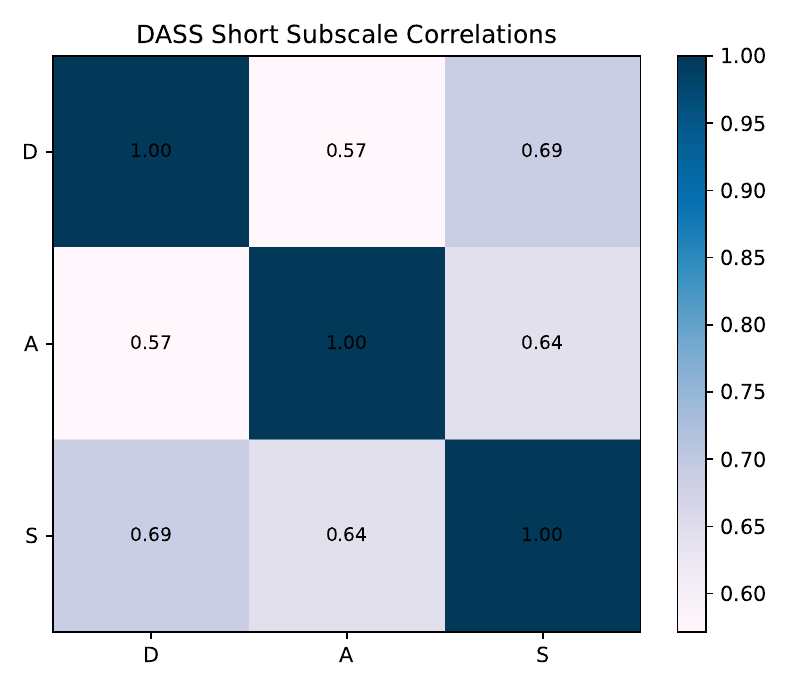}}
	\caption{Structural Similarity of Subscale Correlation Patterns (DASS)}
	\label{fig:re-ma-sim-dass}
\end{figure}

\begin{figure}[ht]
	\centering 
	\subfigure[ ]{
		\label{i-level.sub.1}
		\includegraphics[width=0.4\linewidth]{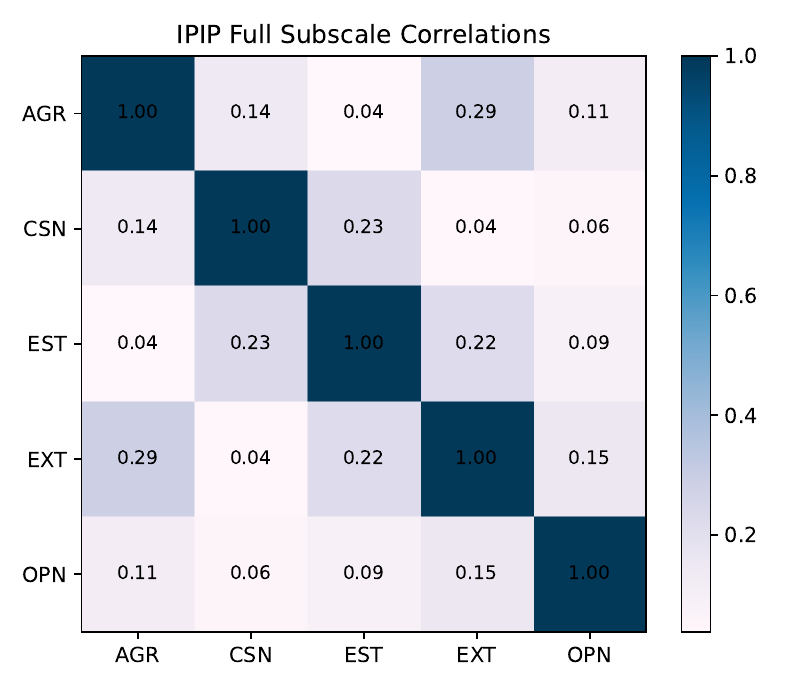}}
	\quad 
	\subfigure[ ]{
		\label{i-level.sub.2}
		\includegraphics[width=0.4\linewidth]{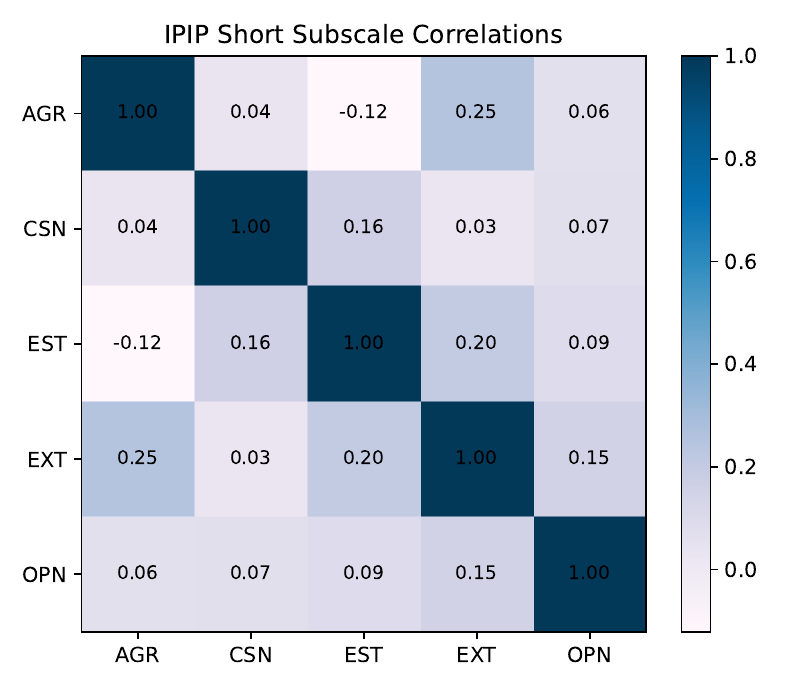}}
	\caption{Structural Similarity of Subscale Correlation Patterns (IPIP)}
	\label{fig:re-ma-sim-ipip}
\end{figure}

\begin{figure}[ht]
	\centering 
	\subfigure[ ]{
		\label{e-level.sub.1}
		\includegraphics[width=0.4\linewidth]{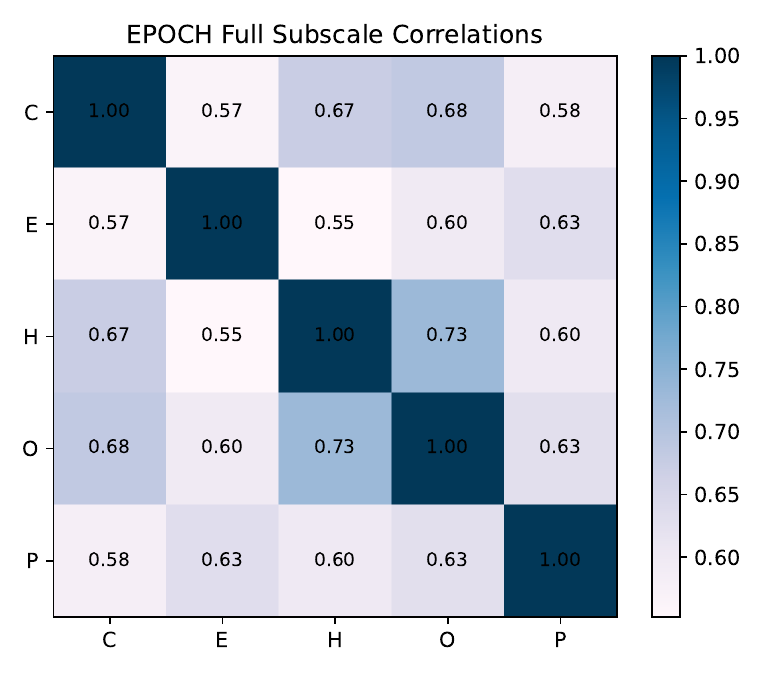}}
	\quad 
	\subfigure[ ]{
		\label{e-level.sub.2}
		\includegraphics[width=0.4\linewidth]{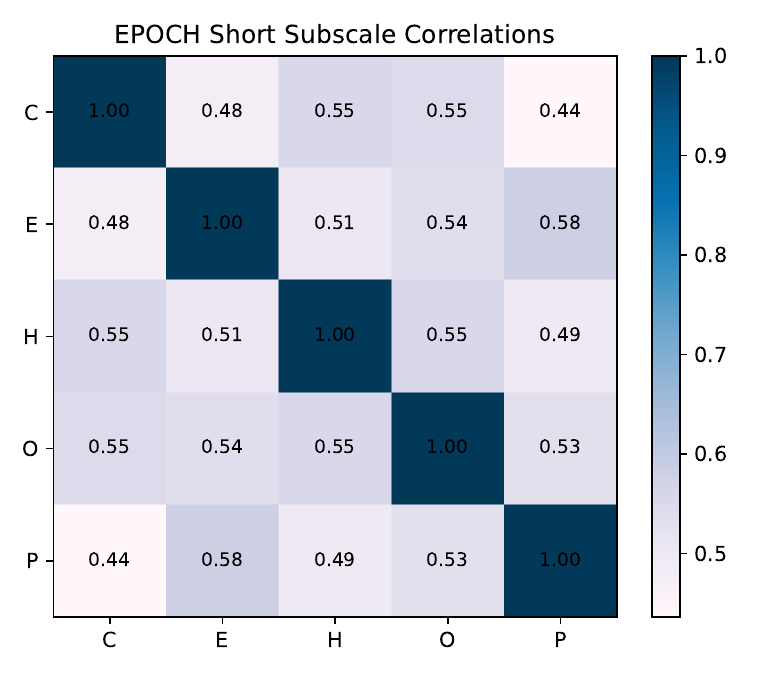}}
	\caption{Structural Similarity of Subscale Correlation Patterns (EPOCH-CN)}
	\label{fig:re-ma-sim-epoch}
\end{figure}

\paragraph{Structural Similarity of Subscale Correlation Patterns (Full vs. Short Forms)}
To examine whether the proposed semantic framework preserves the internal relational structure of psychological constructs, we compared the subscale-level correlation patterns of the full and shortened versions of each questionnaire. For each scale, Pearson correlation matrices were computed among subscale scores derived from the full form and the short form, respectively. Heatmaps of these matrices are presented in Figures \ref{fig:re-ma-sim-dass} (DASS), \ref{fig:re-ma-sim-ipip} (IPIP), and \ref{fig:re-ma-sim-epoch} (EPOCH-CN).

\subparagraph{DASS}
In the full DASS, correlations among Depression, Anxiety, and Stress were moderate to strong (\textit{r} = .67–.80), reflecting the well-established overlap among negative affective states. The corresponding short-form matrix showed a closely matching structure, with all inter-factor correlations remaining positive and within a comparable range (\textit{r} = .57–.69), despite a modest overall attenuation in correlation magnitudes. Consistent with this visual correspondence, the normalized Frobenius similarity between the two correlation matrices was high (similarity = 0.889), indicating substantial preservation of the original relational structure.

\subparagraph{IPIP}
For the IPIP, the full-form correlation matrix exhibited the characteristic Big Five pattern, with generally low to moderate inter-factor correlations (most |\textit{r}| < .30). The short-form matrix closely mirrored this structure, with inter-factor correlations remaining weak (typically |\textit{r}| $\leq$ .25) and no emergence of spurious strong associations. The overall configuration of trait relationships observed in the full scale was therefore maintained in the shortened version. The normalized Frobenius similarity between the full and short matrices was 0.872, further supporting structural consistency.

\subparagraph{EPOCH-CN}
In the EPOCH-CN measure, the full-form subscale correlations were moderate to high (\textit{r} $\approx$ .55–.73), consistent with the presence of a shared higher-order well-being factor. The short-form matrix displayed a similar relational pattern, with inter-factor correlations ranging approximately from .44 to .58. While correlations were systematically lower in the short form, the relative strength of associations among dimensions remained stable. Structural similarity remained high (similarity = 0.861), despite a small attenuation in inter-factor correlations.

To conclude, across all three instruments, the short forms reproduced the characteristic subscale correlation patterns of their full-length counterparts. Although correlation magnitudes were generally reduced, the relative configuration of relationships among dimensions was preserved, indicating that the proposed semantic-based item selection maintained the internal structural organization of the original scales.

\begin{figure}[ht]
	\centering 
	\subfigure[DASS]{
		\label{fig:re-subscale-sim-dass}
		\includegraphics[width=0.3\linewidth]{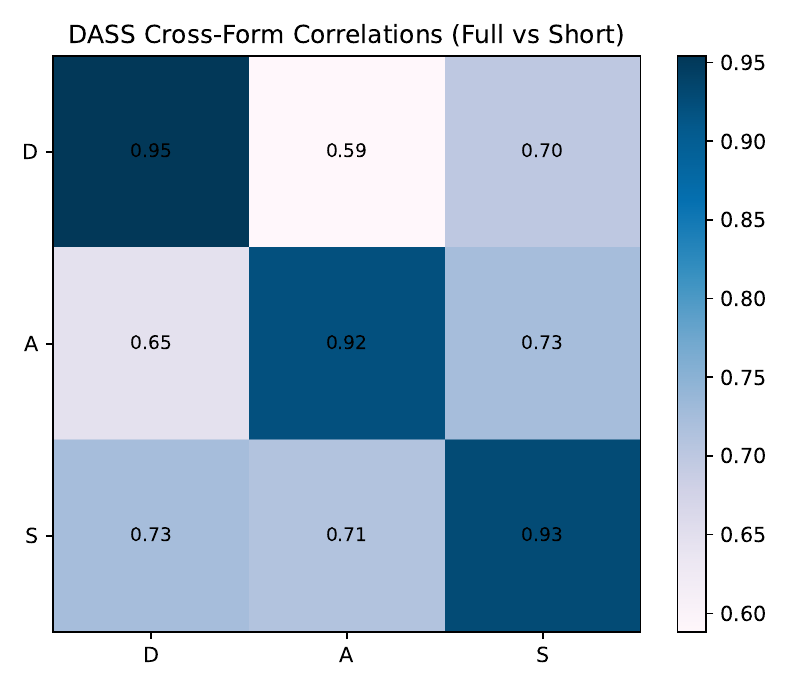}}
	\quad 
	\subfigure[IPIP]{
		\label{fig:re-subscale-sim-ipip}
		\includegraphics[width=0.3\linewidth]{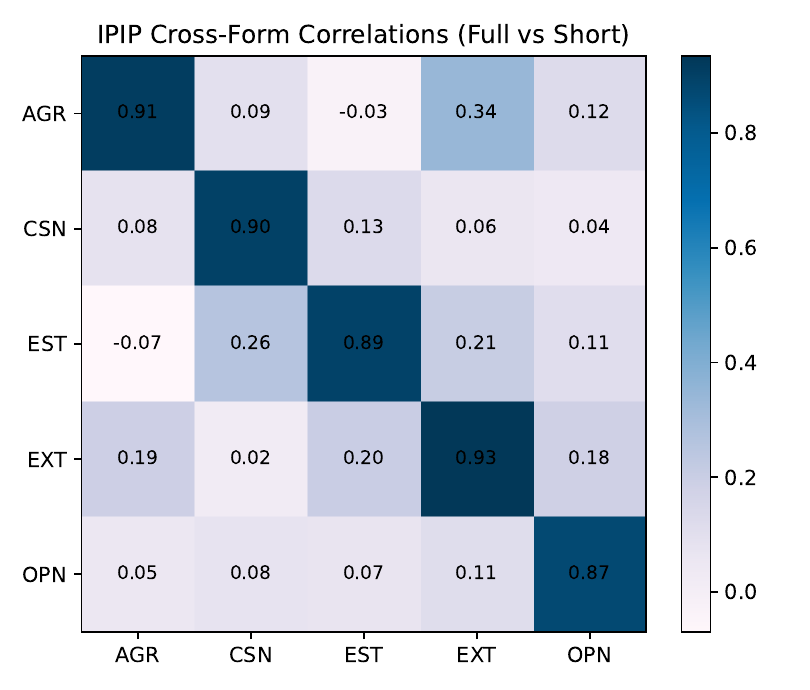}}
	\quad 
	\subfigure[EPOCH-CN]{
		\label{fig:re-subscale-sim-epoch}
		\includegraphics[width=0.3\linewidth]{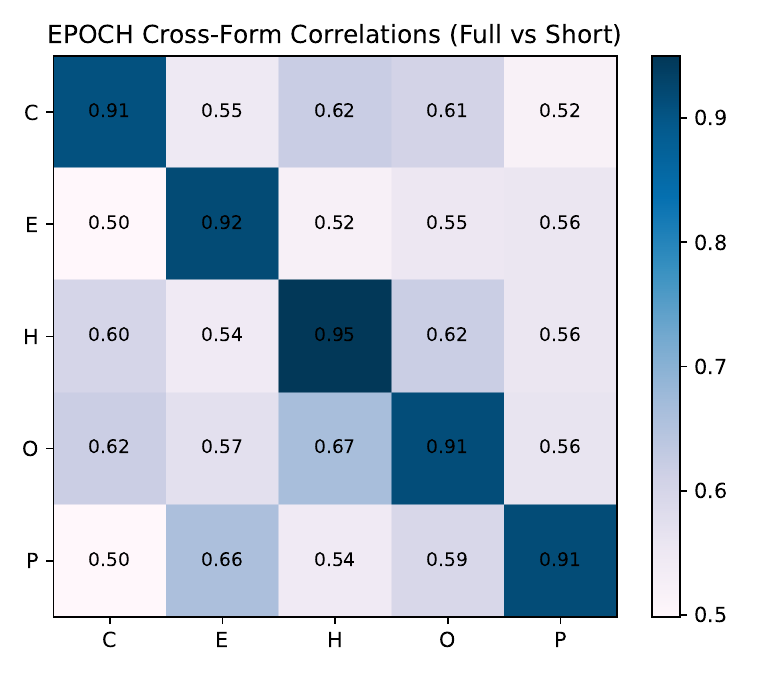}}

	\caption{Cross-Form Subscale Correlations}
	\label{fig:re-subscale-sim}
\end{figure}

\paragraph{Convergent and Discriminant Associations Between Full and Short Forms (Cross-Form Subscale Correlations)}
To assess whether the shortened versions accurately captured the intended constructs of the original scales, we computed cross-form Pearson correlations between subscale scores derived from the full and short forms. For each questionnaire, a k × k cross-correlation matrix was obtained (k as the number of factors, k = 3 for DASS; k = 5 for IPIP and EPOCH-CN), where diagonal elements reflect convergent associations between corresponding dimensions, and off-diagonal elements reflect discriminant associations between non-corresponding dimensions. Heatmaps of these matrices are presented in Figures \ref{fig:re-subscale-sim-dass} (DASS), \ref{fig:re-subscale-sim-ipip} (IPIP), and \ref{fig:re-subscale-sim-epoch} (EPOCH-CN).

\subparagraph{DASS}
For the DASS, strong convergent correlations were observed between corresponding subscales of the full and short forms. Diagonal correlations ranged from approximately \textit{r} = .92 to .95, indicating a high degree of alignment between the two versions for Depression, Anxiety, and Stress. In contrast, off-diagonal correlations were consistently lower (generally \textit{r} < .75), reflecting partial overlap among negative affect dimensions while remaining clearly below the corresponding diagonal values.

\subparagraph{IPIP}
In the IPIP, convergent correlations between full-form and short-form subscales were uniformly high across all five personality traits, with diagonal correlations ranging from approximately \textit{r} = .87 to .93. Off-diagonal correlations were substantially lower, typically falling below .30. This pattern indicates that the shortened version preserved strong trait-specific correspondence with the full scale, while maintaining the relative independence among distinct personality dimensions.

\subparagraph{EPOCH-CN}
For the EPOCH-CN, convergent correlations between full and short forms were similarly strong, with diagonal values ranging from approximately \textit{r} = .91 to .95 across the five well-being dimensions. Off-diagonal correlations were moderate (approximately \textit{r} = .50–.67), consistent with the presence of a shared higher-order well-being factor in both versions of the scale. Importantly, for all dimensions, diagonal correlations exceeded off-diagonal correlations, indicating clear construct alignment between the full and short forms.

Across all three instruments, cross-form correlation analyses demonstrated strong convergent associations between corresponding subscales of the full and short forms, alongside comparatively weaker discriminant associations between non-corresponding dimensions. This pattern indicates that the proposed semantic framework yields shortened scales that closely align with their full-length counterparts at the construct level.

\subsubsection{Semantic Structure Visualization}
\begin{figure}[ht!]
\centering
    \caption{Two-Dimensional Visualization of the Semantic Representation Space for DASS}
    \includegraphics[width=0.9\linewidth]{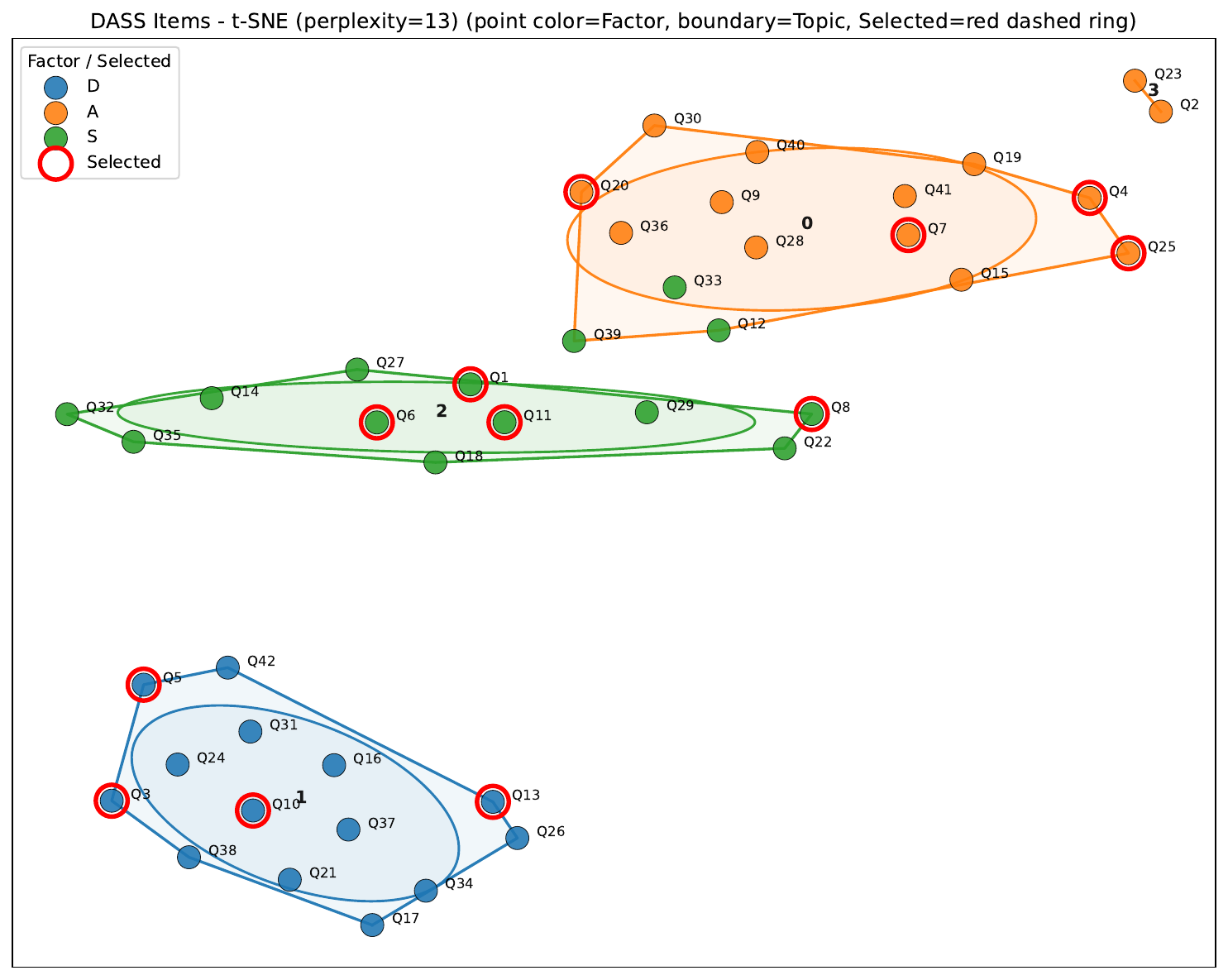}
    \label{fig:re-tsne-dass}
    
    \begin{minipage}{0.85\textwidth}
        \vspace{2mm}
        \footnotesize
        \textit{Note}. Each point represents an item of the scale, and colors indicate the true factor to which the item belongs. The boundary is estimated by the proposed method. Items highlighted with red dashed circles indicate those selected as representative items for the shortened scale.
    \end{minipage}
\end{figure}

\begin{figure}[ht!]
\centering
    \caption{Two-Dimensional Visualization of the Semantic Representation Space (IPIP).}
    \includegraphics[width=0.9\linewidth]{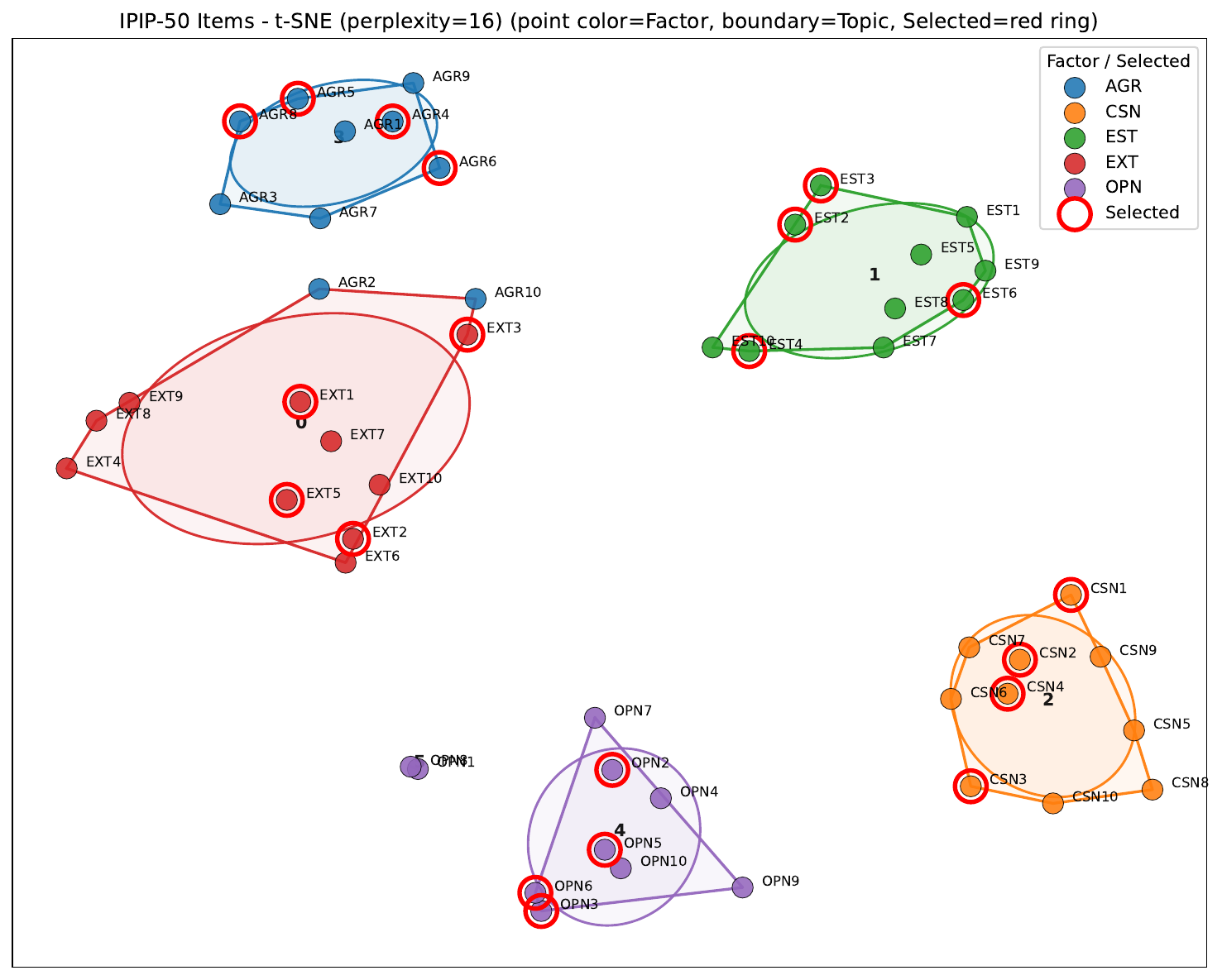}
    \label{fig:re-tsne-ipip}
\end{figure}

\begin{figure}[ht!]
\centering
    \caption{Two-Dimensional Visualization of the Semantic Representation Space (EPOCH-CN).}
    \includegraphics[width=0.9\linewidth]{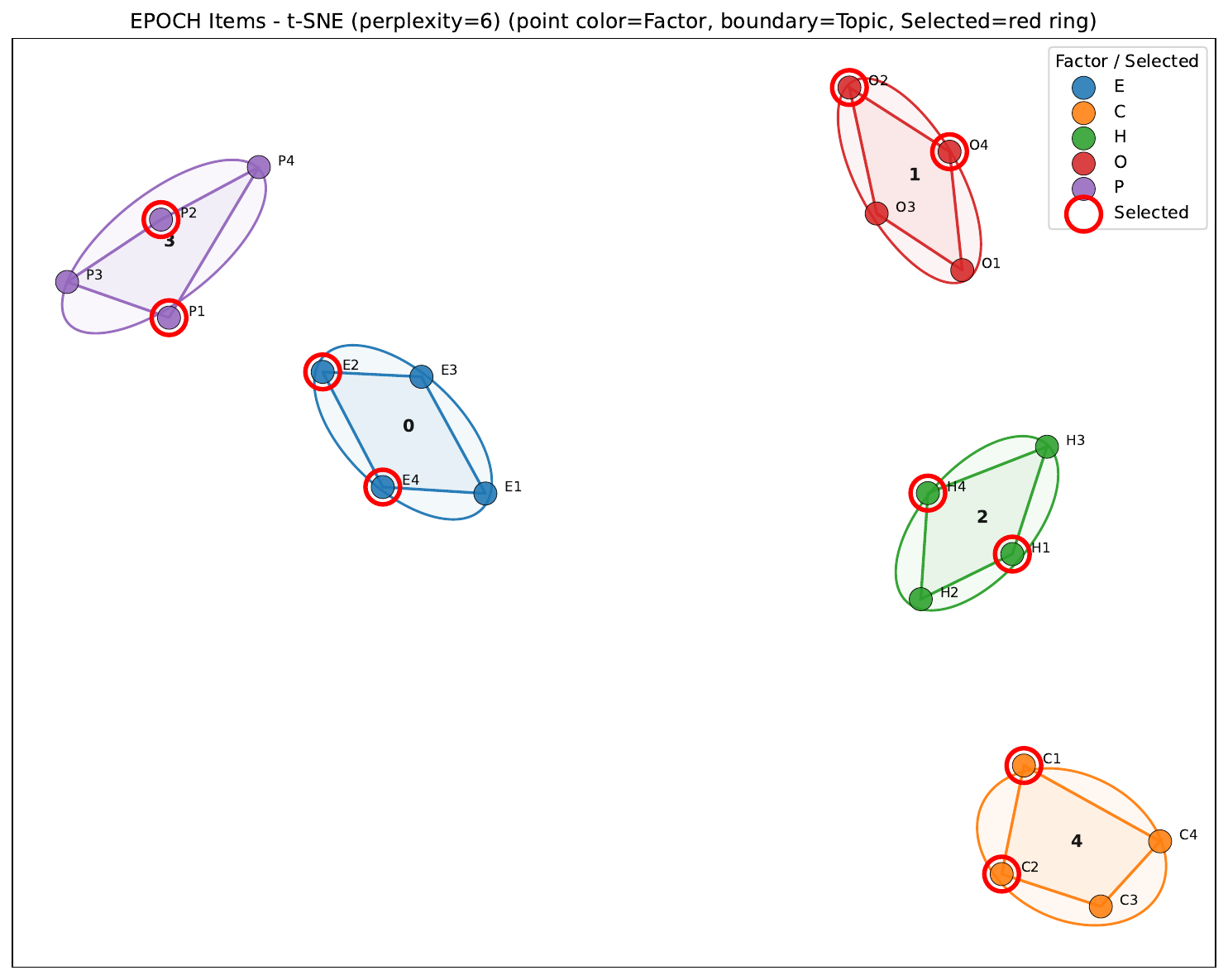}
    \label{fig:re-tsne-epoch}
\end{figure}

Figures \ref{fig:re-tsne-dass},\ref{fig:re-tsne-ipip} and \ref{fig:re-tsne-epoch} present two-dimensional t-SNE visualizations of item embeddings for the DASS, IPIP and EPOCH-CN scales, respectively. In each figure, individual items are colored according to their original theoretical factors, while topic boundaries reflect the clusters derived from the topic modeling procedure. Items selected for the simplified versions are highlighted with red rings.
Across all three scales, the topic modeling results show a high degree of correspondence with the original factor structures, despite the absence of any explicit constraint or setting on the number or composition of factors during clustering.

\paragraph{DASS}
For the DASS scale (Figure \ref{fig:re-tsne-dass}), three major topic clusters emerged and broadly aligned with the Depression (D), Anxiety (A), and Stress (S) factors. Most items associated with each theoretical factor were grouped within the same topic boundary, indicating that the semantic embedding space preserved the original tripartite structure of the scale.
A small number of items appeared near the boundaries between adjacent clusters. For example, several stress-related items were positioned close to the anxiety cluster. Despite that, these boundary cases did not disrupt the overall alignment between topics and factors, but instead reflected localized deviations within an otherwise coherent structure.
The selected items were distributed across all three clusters and tended to occupy central or representative positions within their respective topic regions.

\paragraph{IPIP}
For the IPIP scale (Figure \ref{fig:re-tsne-ipip}), which contains a larger number of items and greater semantic heterogeneity, topic modeling nevertheless produced coherent clusters corresponding to the Big Five dimensions. The majority of items within each topic shared the same factor label (AGR, CSN, EST, EXT, or OPN), and topic boundaries largely aligned with the theoretical structure of the scale.

Some clusters exhibited more elongated or irregular shapes in the embedding space, and a limited number of items were positioned closer to neighboring factor clusters. For instance, a small subset of agreeableness items appeared adjacent to extraversion-related items. Despite these localized deviations, the overall correspondence between topics and factors remained strong.
Selected items were spread across all five dimensions and were typically located near the centers of their respective topic clusters, suggesting that the simplified version retained broad semantic coverage of the original scale.

\paragraph{EPOCH-CN}
For the EPOCH-CN scale (Figure \ref{fig:re-tsne-epoch}), the five topic clusters exhibited a near one-to-one correspondence with the theoretical dimensions of Engagement (E), Perseverance (P), Connectedness (C), Optimism (O), and Happiness (H). Each topic boundary predominantly enclosed items from a single factor, with minimal overlap between clusters.
Compared to DASS and IPIP, the clusters in EPOCH-CN were more compact and clearly separated in the semantic space. Selected items were consistently drawn from within the core regions of each cluster. No topic was dominated by items from multiple factors, and no factor was split across multiple topics.

Taken together, the visualization results demonstrate that the proposed topic modeling method can recover the primary factor structures of established psychological scales while allowing for limited, interpretable deviations at the item level. Across scales with differing numbers of dimensions and degrees of semantic heterogeneity, the selected items consistently reflected the central semantic content of each topic cluster.

\subsubsection{Parameter Setting Guidance and Stability Analysis}

\begin{figure}[ht]
\centering
    \caption{CFI and TLI Trend Graph under Different Number of Items}
    \includegraphics[width=0.75\linewidth]{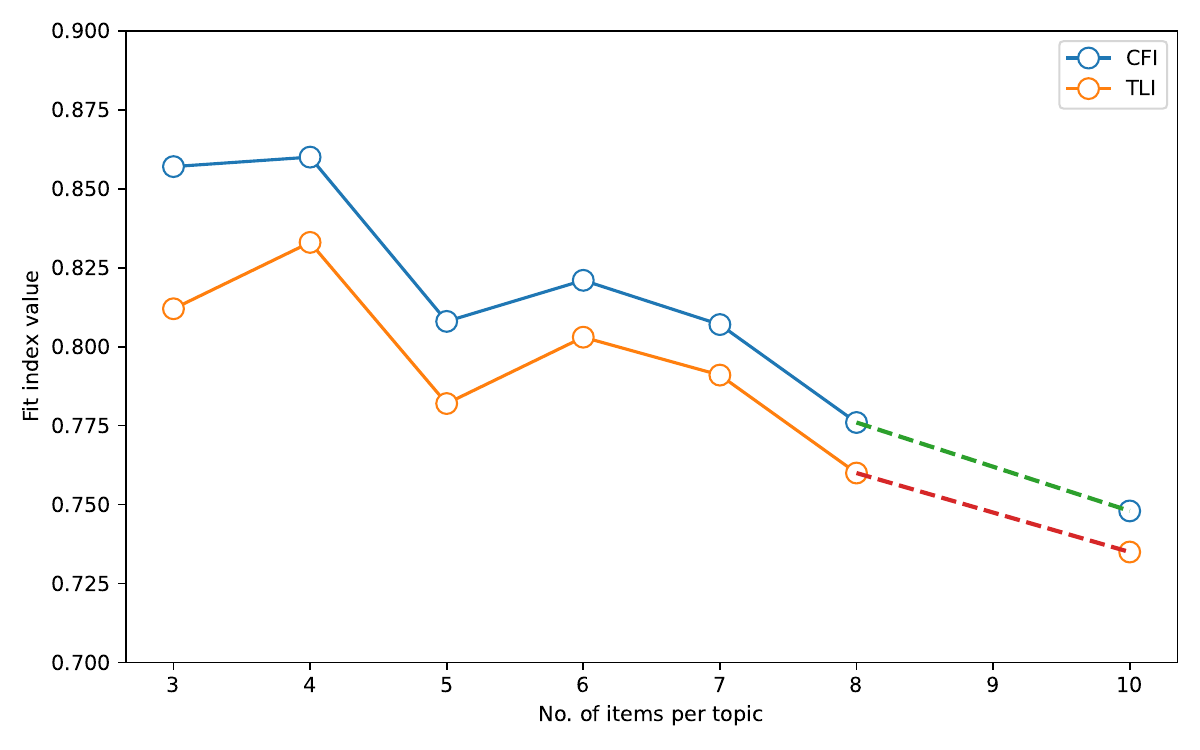}
    \label{fig:cfa-fig-num-of-item}
\end{figure}

\paragraph{Structural Analysis Under Key Parameter Change}
\subparagraph{Number of items retained} We first examined how the number of items retained per semantic topic influences the structural fit of the simplified scale, with the IPIP items fixed to five topics. Table \ref{tab:cfa-num-of-item} and Figure \ref{fig:cfa-fig-num-of-item} report the resulting CFI and TLI values across different items-per-topic settings.
As shown, retaining only three to four items per topic was sufficient to achieve relatively good model fit. Specifically, when three items were retained per topic, CFI = 0.857 and TLI = 0.812; when four items were retained, CFI = 0.860 and TLI = 0.833. Beyond this range, both fit indices exhibited a non-monotonic pattern, characterized by an initial decline, a brief recovery, and a subsequent downward trend as more items were included.

\begin{table}[htbp]
  \centering
  \caption{CFI and TLI Analysis under Different Number of Items}
      \resizebox{0.4\linewidth}{!}{
    \begin{tabular}{ccc}
    \toprule
    \textbf{No. of items per topic} & \textbf{CFI} & \textbf{TLI} \\
    \midrule
    3     & 0.857 & 0.812 \\
    4     & 0.860  & 0.833 \\
    5     & 0.808 & 0.782 \\
    6     & 0.821 & 0.803 \\
    7     & 0.807 & 0.791 \\
    8     & 0.776 & 0.760  \\
    9     & -     & - \\
    10    & 0.748 & 0.735 \\
    \bottomrule
    \end{tabular}}%
  \label{tab:cfa-num-of-item}%
\end{table}%

This pattern reflects a trade-off between information sufficiency and semantic focus during scale simplification: when too few items are retained within each topic (e.g., one or two items), the intended measurement model cannot be reliably identified, and model estimation even failed to converge, indicating insufficient information to support a stable measurement model. Conversely, as additional items are incorporated, the apparent increase in information is accompanied by the inclusion of items that are semantically more peripheral to the core topic, thereby reducing semantic coherence and structural clarity. This effect becomes observable as early as selecting five items per topic.
Although a temporary improvement in fit was observed when six items per topic were retained, which is likely due to increased item information, both CFI and TLI subsequently declined as more items were added. Overall, these results indicate that scale simplification does not benefit monotonically from retaining more items. Instead, optimal performance emerges from balancing semantic representativeness with a sufficient, but not excessive, number of items per topic.

For practical parameter settings, researchers may therefore consider starting with relatively aggressive simplification strategies, that is, retaining only the highest-probability items within each topic (approximately 30\%–40\% of the original scale length), and then refining the resulting short form through standard psychometric evaluation.

\subparagraph{Number of topics}
We next examined the effect of varying the number of semantic topics (by setting the \textit{nr\_topics} parameter) on the structural fit of the simplified scale using the IPIP dataset. To isolate the impact of topic granularity, the total number of retained items was fixed at 20 across all conditions, with the number of items allocated to each topic determined proportionally based on topic size.

\begin{figure}[ht]
\centering
    \caption{CFI and TLI Trend Graph under Different Number of Topics}
    \includegraphics[width=0.75\linewidth]{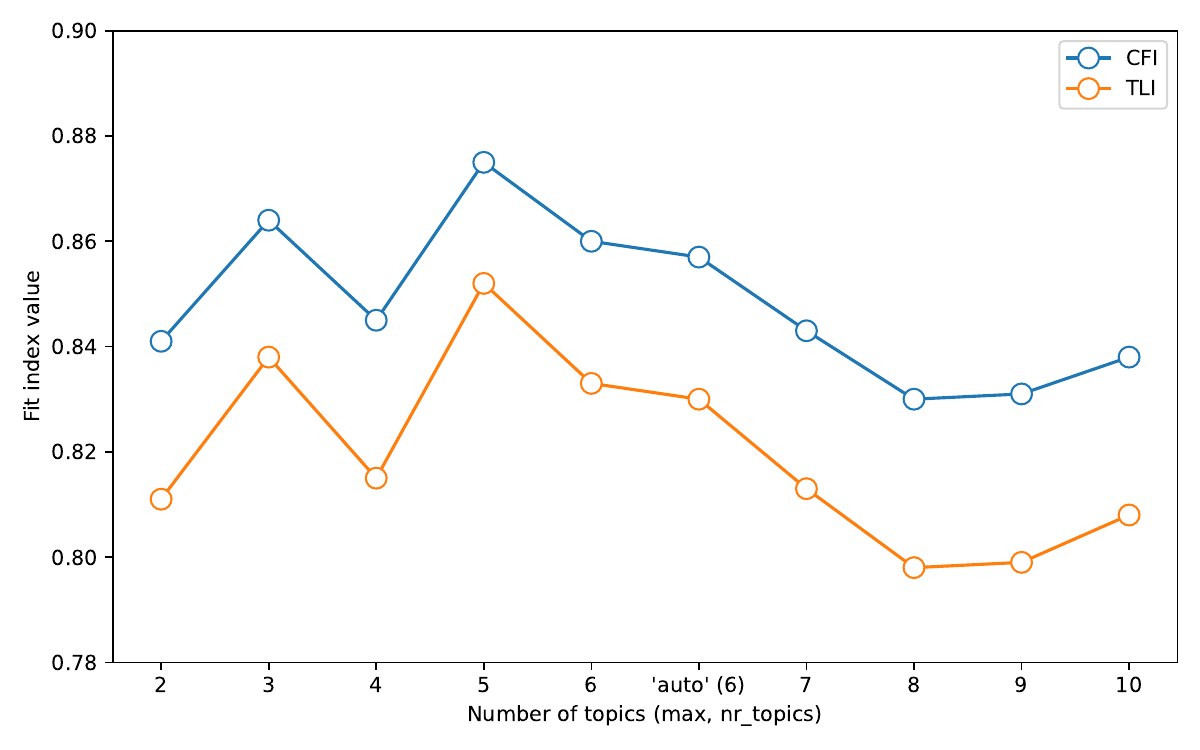}
    \label{fig:cfa-fig-num-of-topic}
\end{figure}

As shown in Table \ref{tab:cfa-num-of-topic} and Figure \ref{fig:cfa-fig-num-of-topic}, the results largely align with our prior expectation that ``model fit is optimized when the number of semantic topics approximates the scale’s underlying factor structure’’. In the present case, the best overall fit was obtained when topics = 5 (CFI = 0.875, TLI = 0.852). In contrast, both fewer and more fine-grained topic settings were associated with reduced model fit.

\begin{table}[htbp]
  \centering
  \caption{CFI and TLI Analysis under Different Number of Topics}
      \resizebox{0.4\linewidth}{!}{
    \begin{tabular}{ccc}
    \toprule
    \textbf{Topics (nr\_topic)} & \textbf{CFI} & \multicolumn{1}{l}{\textbf{TLI}} \\
    \midrule
    2     & 0.841 & 0.811 \\
    3     & 0.864 & 0.838 \\
    4     & 0.845 & 0.815 \\
    5     & 0.875 & 0.852 \\
    6     & 0.860  & 0.833 \\
    `auto' (6) & 0.857 & 0.830 \\
    7     & 0.843 & 0.813 \\
    8     & 0.830  & 0.798 \\
    9     & 0.831 & 0.799 \\
    10    & 0.838 & 0.808 \\
    \bottomrule
    \end{tabular}}%
  \label{tab:cfa-num-of-topic}%
\end{table}%

A notable deviation from this general trend was observed at topics = 4, where CFI = 0.845, TLI = 0.815, both were substantially lower than in adjacent conditions. Further inspection revealed that this decrease was not due to random variability, but rather likely to be the changes in the semantic clustering structure. When topics = 3, Extraversion (EXT) and Agreeableness (AGR) were actually grouped into a single topic, as were Conscientiousness (CSN) and Openness (OPN), while Emotional Stability (EST) formed a separate topic. Despite the reduced number of topics, the resulting item selection remained relatively balanced in terms of semantic coverage and item quality, yielding comparatively strong model fit.
In contrast, under the topics = 4 condition, semantically atypical OPN items (e.g., OPN1 and OPN8) were separated into an additional topic distinct from the main OPN cluster, whose items primarily emphasize ideas and imagination. As a consequence, the final selection step was forced to retain at least one item from this extra topic such as OPN8 (``I use difficult words’’), thereby constraining the selection of more representative items from other topics and ultimately degrading overall model fit.
When the automatic topic-determination mode was used (\textit{nr\_topics} = `auto'), the resulting model also achieved reasonably strong fit (CFI = 0.857, TLI = 0.830), comparable to manually specified topic settings in the same range.

For practical parameter settings, when researchers have prior knowledge of the approximate number of latent factors underlying a scale, it is suggested that setting \textit{nr\_topics} equal to that number or slightly higher (i.e., n or n + 1), which should offer a pragmatic balance between semantic differentiation and structural stability.

\paragraph{Stability Analysis Under Parameter Perturbation}

\begin{table}[htbp]
  \centering
  \caption{Simplification Result and Jaccard Value under Key Parameter Perturbation (DASS)}
    \resizebox{1\linewidth}{!}{
    \begin{tabular}{m{8em}<{\centering}m{9em}<{\centering}m{14em}<{\centering}m{6em}<{\centering}m{5em}<{\centering}m{5em}}
    \toprule
    \multicolumn{1}{c}{Parameter} & Setting & Reduction result & Jaccard & Kept  & Changed \\
    \midrule
    default & n\_neighbors = 3 min\_cluster\_size=2 min\_samples = 1 & \{1,3,4,5,6,7,8,10,11,13,20,25\} &       &       &  \\
          &       &       &       &       &  \\
    n\_neighbors & 2     & \{1,3,4,5,7,8,9,10,12,13,14,19\} & 0.500 & 8     & 4 \\
    (UMAP) & 10    & \{1,2,3,4,5,6,7,8,10,13,16,17\} & 0.600 & 9     & 3 \\
    min\_cluster\_size & 4     & \{1,3,4,6,7,9,10,11,15,16,17,18\} & 0.412 & 7     & 5 \\
    (HDBSCAN) & 6     & \{1,3,4,5,6,7,10,11,12,15,16,19\} & 0.500 & 8     & 4 \\
    min\_samples & 2     & \{1,3,4,6,7,8,9,10,11,15,17,21\} & 0.500 & 8     & 4 \\
    (HDBSCAN) & 3     & \{1,4,6,9,11,15,19,21,24,27,31,38\} & 0.200 & 4     & 8 \\
    \bottomrule
    \end{tabular}}%
  \label{tab:para-stable-dass}%
\end{table}%

\begin{table}[htbp]
  \centering
  \caption{Simplification Result and Jaccard Value under Key Parameter Perturbation (IPIP)}
    \resizebox{1\linewidth}{!}{
    \begin{tabular}{m{8em}<{\centering}m{9em}<{\centering}m{14em}<{\centering}m{6em}<{\centering}m{5em}<{\centering}m{5em}}
    \toprule
    \multicolumn{1}{c}{Parameter} & Setting & Reduction result & Jaccard & Kept  & Changed \\
    \midrule
    default & n\_neighbors = 3 min\_cluster\_size=2 min\_samples = 1 & \{1,2,3,5,12,13,14,16,24,25,26,28,  31,32,33,34,42,43,45,46\} &       &       &  \\
          &       &       &       &       &  \\
    n\_neighbors & 2     & \{1,2,3,6,7,11,12,13,14,23,24,25,   32,33,34,35,42,43,44,46\} & 0.538 & 14    & 6 \\
    (UMAP) & 10    & \{1,2,3,11,12,13,14,15,16,17,20,21,  22,31,32,33,34,41,42,43\} & 0.481 & 13    & 7 \\
    min\_cluster\_size & 4     & \{1,2,3,4,11,14,15,16,21,23,24,25,  31,32,33,34,41,42,43,44\} & 0.481 & 13    & 7 \\
    (HDBSCAN) & 6     & \{1,2,3,4,5,6,8,9,11,14,15,16,  31,32,34,36,41,42,43,44\} & 0.379 & 11    & 9 \\
    min\_samples & 2     & \{3,5,6,7,11,14,15,16,24,26,27,28,  31,32,34,35,42,44,45,47\} & 0.429 & 12    & 8 \\
    (HDBSCAN) & 3     & \{3,5,6,7,8,9,10,11,14,15,16,21,  32,34,36,37,41,42,43,44\} & 0.250 & 8     & 12 \\
    \bottomrule
    \end{tabular}}%
  \label{tab:para-stable-ipip}%
\end{table}%

\begin{table}[htbp]
  \centering
  \caption{Simplification Result and Jaccard Value under Key Parameter Perturbation (EPOCH-CN)}
    \resizebox{1\linewidth}{!}{
    \begin{tabular}{m{8em}<{\centering}m{9em}<{\centering}m{14em}<{\centering}m{6em}<{\centering}m{5em}<{\centering}m{5em}}
    \toprule
    \multicolumn{1}{c}{Parameter} & Setting & Reduction result & Jaccard & Kept  & Changed \\
    \midrule
    default & n\_neighbors = 3 min\_cluster\_size=2 min\_samples = 1 & \{1,2,4,7,9,10,12,13,18,20\} &       &       &  \\
          &       &       &       &       &  \\
    n\_neighbors & 2     & \{2,4,5,6,7,9,10,14,15,18\} & 0.429 & 6     & 4 \\
    (UMAP) & 10    & \{1,2,4,5,6,7,9,10,13,18\} & 0.667 & 8     & 2 \\
    min\_cluster\_size & 4     & \{1,2,3,4,5,6,7,9,10,13\} & 0.538 & 7    & 3 \\
    (HDBSCAN) & 6     & Unassigned* & /     & /     & / \\
    min\_samples & 2     & \{4,7,10,11,13,14,15,17,19,20\} & 0.333 & 5     & 5 \\
    (HDBSCAN) & 3     & \{4,5,10,11,13,14,17,18,19,20\} & 0.333 & 5     & 5 \\
    \bottomrule
    \multicolumn{6}{l}{*: We failed to obtain the reduction results due to all items under this setting are judged as ``-1'' (noisy points).}
    \end{tabular}}%
  \label{tab:para-stable-epoch}%
\end{table}%

The simplification result under key parameter perturbations are shown in Tables \ref{tab:para-stable-dass}, \ref{tab:para-stable-ipip} and \ref{tab:para-stable-epoch}.
Across the three scales, representative item selection showed systematic but interpretable variation under targeted hyperparameter perturbations. 

For DASS, overlap with the default selection remained moderate under most perturbations (Jaccard = 0.41-0.60; 3-5 items replaced out of 12), indicating relatively stable core representatives. Notably, stability decreased under the most stringent core-point setting (\textit{min\_samples} = 3; Jaccard = 0.20; 8 items replaced), suggesting that extreme density requirements can substantially shift representative items. 

For IPIP, overlap decreased more noticeably as perturbations became stronger (Jaccard = 0.25-0.54; 6-12 items replaced out of 20) in all three parameter changes, reflecting greater semantic heterogeneity and multiple near-equivalent candidates within broader personality content. 

For EPOCH-CN, \textit{min\_samples} seemed to have a stronger impact than the other two parameters when setting to a higher value, giving lower Jaccard incides (Jaccard = 0.333, 5 items replaced out of 10 for both situations). Notably, when \textit{min\_cluster\_size} = 6, the clustering procedure returned all items as ``unassigned''. This outcome is expected since the specified \textit{min\_cluster\_size} exceeded the typical number of items per factor in EPOCH-CN (five factors with four items each). The fact that no larger merged clusters were formed suggests that items in EPOCH-CN were well-designed, where items belonging to different factors were sufficiently well-separated in semantic space.

Overall, a larger departure from the default configuration tended to yield lower overlap with the default item set, consistent with the expectation that increasingly conservative or aggressive neighborhood and density settings alter topic granularity and core-point identification.

\newpage

\section{Discussion}
\subsection{Technical Foundation and Highlights}

The present framework addresses a practical gap in scale construction and simplification. Traditional psychometric and machine learning-assisted reduction methods remain indispensable for establishing measurement quality, but they usually require respondent data before item selection can be performed. This dependence can be costly in early-stage scale development, cross-cultural adaptation, or other settings where large validation samples are not yet available. At the same time, questionnaire items are not arbitrary indicators: they are linguistic expressions of psychological constructs. Their wording often contains semantic redundancy, conceptual overlap, and narrower subthemes that can be inspected before response data are collected. Our framework formalizes this often implicit stage of scale development by using item wording to generate a content-informed semantic pre-structure and to narrow candidate item pools before response-based modeling.

The technical basis of the framework is semantic embedding. This treatment is consistent with recent work showing that natural language processing can transform text into analyzable representations for behavioral science \parencite{feuerriegel2025using}, and with vector-space models of semantic representation, which represent meaning as high-dimensional numerical vectors learned from patterns of language use \parencite{gunther2019vector}. 
In this study, each questionnaire item is treated as a short natural-language statement and encoded by a pretrained language model into a dense numerical vector. Such vectors are not simple word counts, but contextual representations learned from large-scale language data, where words and sentences that appear in similar linguistic contexts tend to acquire similar representations. As a result, two items with similar meanings are expected to occupy nearby regions in the embedding space, even when they do not share identical surface words. This property makes embeddings suitable for representing item-level semantic similarity and for examining whether questionnaire items form coherent semantic neighborhoods.

This representation also explains why dimensionality reduction and clustering are appropriate subsequent steps. The original embedding space is high-dimensional and difficult to inspect directly. Dimensionality-reduction methods provide a lower-dimensional approximation that preserves major neighborhood relations among items, making the semantic organization more tractable for visualization and clustering. Density-based clustering then identifies groups of items that occupy locally dense regions of this semantic space, without requiring the number of clusters to be fixed in advance. In the context of scale simplification, these clusters are interpreted cautiously as semantic groupings or semantic ``factor-like'' structures, rather than as psychometric latent factors themselves.

A further contribution of the framework is that it links geometric clustering with interpretable topic refinement. After semantic clusters are identified, cluster-level keywords are extracted from the original item texts using \textit{c-TF-IDF}, making each cluster’s thematic core inspectable to researchers. These topic representations also support a secondary consolidation step, in which closely related clusters can be merged when they reflect overlapping lexical themes. Thus, the framework does not rely on embedding-space clustering alone, but adds an explicit topic-based layer for interpretation and refinement. Representative items are then selected according to their semantic membership within each refined topic, analogous in spirit to retaining high-loading items in factor-analytic reduction, but derived from semantic organization rather than response covariance.

It should be noted that the framework is not intended to replace psychometric validation. Semantic proximity does not guarantee equivalent response processes, item difficulty, reliability, or criterion validity. Instead, the proposed method functions as a response-free front-end: it produces an explicit, reproducible, and interpretable candidate structure that can guide subsequent empirical evaluation. Candidate short forms are therefore assessed using conventional psychometric evidence, including model fit, factor loadings, reliability, score correspondence, and stability across parameter settings. Across the case studies, this integration supports transparent early-stage item selection while remaining compatible with established validation workflows.

Overall, the main contribution of this work is threefold. First, it introduces a response-free topic-modeling framework that treats item texts as analyzable semantic data rather than only as observed indicators of response models. Second, it integrates contextual embeddings, dimensionality reduction, density-based clustering, topic labeling, and representative-item selection into a unified and inspectable pipeline for scale simplification. Third, it demonstrates that semantic pre-structures can guide efficient scale reduction while preserving the central role of downstream psychometric validation.

\subsection{Result and Performance Discussion Through Case Studies}
Across the three case studies, a consistent lesson is that the proposed text-first simplification method performs best when the target scale originally exhibits high within-factor semantic coherence and clear semantic boundaries between factors. EPOCH-CN provides the clearest illustration. In the semantic space, clusters were distinctly separated with minimal apparent mixing across factors (see Figure \ref{fig:re-tsne-epoch}), suggesting that the item pool contains well-differentiated content themes that align closely with the intended subscales. This strong semantic separability translated into robust measurement performance even under an aggressive length constraint (two items per factor). In the short-form CFA, standardized loadings remained moderate-to-strong across factors (approximately .58–.87; e.g., C items .58–.72, H items .83–.87, E items .65–.78, P items .66–.72, O items .61–.78; see Table \ref{tab:re-fact-load-epoch}). Collectively, these results suggest that when a scale’s content structure is linguistically well-partitioned, semantic clustering and topic modeling can serve as an effective front-end for generating short-form candidates that remain psychometrically defensible after validation.

EPOCH-CN also demonstrates that semantic alignment can help preserve relational structure among dimensions, not merely within-dimension measurement quality. The full-scale's subscale correlation matrix showed a coherent pattern of moderate inter-factor associations (e.g., .55–.73 among several pairs), and the short form reproduced a similar rank-order pattern, albeit with expected attenuation due to reduced item counts (generally .44–.58; see Figure \ref{fig:re-ma-sim-epoch}). When comparing off-diagonal entries, the correspondence between the full and short inter-factor correlation patterns was substantial (Spearman $\rho$ $\approx$ .77; Pearson \textit{r} $\approx$ .63; MAD $\approx$ .10; see Figure \ref{fig:re-subscale-sim-epoch}), indicating that the short form retained much of the structural ``signature'' of the original scale even under severe compression. Practically, this supports the view that the proposed text-driven item selection, when followed by standard psychometric checks, can preserve both factor-level measurement and between-factor network structure when the construct domains are semantically distinct and consistently expressed in item wording.

At the same time, IPIP provides an intermediate case between EPOCH-CN and DASS  (state later in this section). Unlike EPOCH-CN, the Big Five domains are broad personality constructs that contain multiple narrower facets. For example, Conscientiousness includes orderliness, preparedness, carefulness, and attention to detail, whereas Openness/Intellect includes imagination, idea generation, intellectual curiosity, and abstract reasoning. In this setting, a semantic front-end may select items that are content-representative of different facets within the same broad domain, but these items may not always form a highly homogeneous short subscale in response data. 

The lower loadings and comparatively weaker item-level indices of CSN3 and OPN2 illustrate this trade-off. 
CSN3 (``I pay attention to details'') captures a carefulness or attention-to-detail facet of Conscientiousness, whereas other retained Conscientiousness items more strongly emphasize preparedness, organization, and orderliness. Similarly, OPN2 (``I easily understand abstract ideas'') reflects an intellect-oriented aspect of Openness, whereas several other retained Openness items emphasize imagination and idea generation. These findings suggest that semantic representativeness and psychometric homogeneity are related but not identical selection criteria. For broad, facet-rich constructs, semantic item selection may preserve content coverage while producing lower within-subscale homogeneity in very short forms, making post hoc psychometric screening especially important.
Additional comparison with the full IPIP further contextualized this pattern: as reported in the IPIP Results section, some weaker item-level patterns were already present in the original scale, suggesting that the short-form results partly reflect characteristics of the original item pool and the present sample rather than being solely attributable to the semantic reduction procedure.

Finally, the case studies clarify that topic–factor alignment is not guaranteed, especially when constructs are inherently ``sticky'' in language (i.e., overlapping semantics across theoretical dimensions). DASS is informative here: the three subscales (Depression, Anxiety, Stress) are known to be interrelated, and the item pool contains wording that naturally blurs boundaries between anxious arousal and stress-related tension. This semantic overlap can manifest as less stable topic–factor correspondences and localized ``misassignments'' that are arguably conceptually plausible rather than purely algorithmic errors. 
For example, three items theoretically keyed to Stress were grouped with Anxiety by the model, with all items emphasized nervous activation and agitation: ``I felt that I was using a lot of nervous energy'' (Q12), ``I was in a state of nervous tension'' (Q33), and ``I found myself getting agitated'' (Q39) (Figure \ref{fig:re-tsne-dass}). From a construct perspective, these statements contain strong anxious-arousal connotations, making their placement near anxiety-themed content unsurprising. In other words, the ``misclassification'' may not only demonstrate one limitation of the model, but also reflect the lexical and phenomenological proximity of stress-related tension to anxiety rather than merely a failure of the pipeline.

Conversely, the DASS visualization also shows that the pipeline may separate items into distinct topics when they share a narrow, concrete semantic theme that is not central to the broader factor narrative (Figure \ref{fig:re-tsne-dass}). Two Anxiety items, ``I was aware of dryness of my mouth'' (Q2) and ``I had difficulty in swallowing'' (Q23), were isolated into a separate topic. This split is defensible semantically: both items describe specific somatic sensations, and their co-occurrence as a ``bodily symptom'' micro-theme is linguistically coherent. Whether such separation is desirable depends on one’s measurement goal: it may improve content coverage transparency (by highlighting a somatic-symptom subtheme), but it may also risk over-fragmentation if the target is to reproduce a pre-specified theoretical factor model with minimal deviation. 

Overall, these case studies suggest several boundary conditions for the proposed framework. The method is likely to be most useful when a scale contains item groups that are semantically coherent within dimensions and reasonably distinct across dimensions. In such cases, as illustrated by EPOCH-CN, semantic clustering and topic modeling can provide a relatively direct front-end for identifying candidate short forms that remain psychometrically defensible after validation.
The method should be used more cautiously when constructs are broad and facet-rich, as in IPIP, because semantically representative items may preserve content coverage at the cost of lower internal homogeneity in very short subscales. It should also be used cautiously when theoretical factors are semantically intertwined, when items contain narrow symptom-like or wording-specific subthemes, or when a construct is defined more by response patterns than by clearly separable item content. 

In such cases, as illustrated by IPIP and DASS, deviations between semantic topics and theoretical factors should not be treated simply as algorithmic errors or as replacements for psychometric evidence. Instead, they can serve as diagnostic information, highlighting where item wording, construct breadth, facet coverage, or content granularity may require closer theoretical and empirical inspection. To conclude, the framework is best understood as a response-free semantic front-end for candidate generation and item-pool diagnosis, with final scale decisions remaining dependent on conventional validation evidence. A conclusion table of the usage and boundary condition can be found in Table \ref{tab:boundary-conditions}.

\begin{table}[htbp]
	\centering
	\caption{Boundary conditions for applying the proposed semantic front-end}
	\label{tab:boundary-conditions}
	\begin{tabular}{p{0.34\linewidth}p{0.27\linewidth}p{0.34\linewidth}}
		\toprule
		\textbf{Scale or item-pool condition} & \textbf{Expected suitability} & \textbf{Interpretive implication} \\
		\midrule
		High within-factor semantic coherence and clear between-factor boundaries (e.g., EPOCH-CN) & Higher suitability & Semantic topics are more likely to align with established dimensions and support direct candidate short-form generation. \\
		Broad, facet-rich constructs with heterogeneous content within each dimension (e.g., IPIP) & Moderate suitability, with caution & Semantic selection may preserve content coverage, but retained items may show lower internal homogeneity in very short subscales. \\
		Semantically overlapping or ``sticky'' constructs (e.g., DASS) & Moderate suitability, with caution & Topic--factor deviations may reveal meaningful construct overlap rather than simple algorithmic error. \\
		Narrow wording-specific or symptom-specific subthemes (e.g., somatic-symptom items in DASS) & Useful for diagnosis, but may over-fragment & The method can highlight content granularity, but researchers should decide whether such subthemes are theoretically central or peripheral. \\
		Constructs primarily defined by response processes, item difficulty, or external criteria rather than item meaning & Lower suitability as a stand-alone guide & Semantic structure should be treated only as preliminary content evidence and must be supplemented by psychometric validation. \\
		\bottomrule
	\end{tabular}
\end{table}

\subsection{Practical Guidance for Using the Framework}

\paragraph{When to use the framework} The proposed text-first pipeline is particularly suitable when researchers (a) need a principled short form to reduce participant burden, (b) are adapting or revising an existing scale for a new context, or (c) wish to generate an initial structural hypothesis for item pools where response data are limited or expensive to obtain. It can also serve as a pre-registration-friendly procedure for defining candidate short forms before examining response patterns, thereby reducing researcher degrees of freedom at the item selection stage.

\paragraph{How to set expectations} The pipeline outputs candidate short forms whose adequacy should be established empirically. A minimum validation package should include: (1) CFA model fit and factor loadings for the short form, (2) reliability evidence (e.g., internal consistency) for each dimension, and (3) correspondence between full- and short-form scores and/or inter-factor correlation patterns (as shown in Figures \ref{fig:re-ma-sim-dass}-\ref{fig:re-subscale-sim}). Stability checks across reasonable parameter perturbations can further strengthen confidence that selected items are not artifacts of a single configuration.

\paragraph{How to diagnose issues} When semantic clusters do not align with theoretical factors, researchers should (a) inspect whether factor definitions overlap linguistically, (b) check whether reversed items or wording style create ``method clusters'', and (c) consider increasing the number of retained items for factors that show high semantic diversity. If the short form shows acceptable score correspondence but weaker CFA fit, the likely implication is that semantic coverage is preserved but the measurement model is underspecified with too few indicators; conversely, if CFA fit is acceptable but score correspondence is degraded, item selection may have narrowed the construct content too aggressively. In either case, the pipeline’s transparency facilitates targeted adjustments: adding representatives from under-covered semantic subthemes or replacing items that dominate a cluster due to stylistic rather than substantive features.

\subsection{Limitations and Boundary Conditions}
\paragraph{Dependence on language representations} The semantic space induced by an embedding model inevitably reflects its training data and inductive biases. As a result, topic solutions and the consequently selected item representatives may vary across embedding choices, domains, and languages. 
In the present study, we partially address this concern by applying our framework to both English (DASS and IPIP) and Chinese (EPOCH-CN) scales with data collected from the corresponding regions, providing initial evidence of cross-language and cross-cultural robustness. Nevertheless, this evaluation remains limited in scope. Future work should more systematically examine multilingual generalizability, including broader language coverage, alternative embedding models, and formal measurement invariance designs.

\paragraph{Semantic similarity as non-sole measurement equivalence} Items that are close in semantic space can still differ in response processes (e.g., difficulty, extremity, or framing). This is particularly relevant when surface wording introduces strong ``method cues'' (e.g., explicit negations). In the present work, DASS and EPOCH-CN did not involve reverse-keyed items in the administered versions; for IPIP, however, as a light preprocessing step, we rewrote reverse-keyed items into the same scoring direction, which also reduced salient negation markers and harmonized surface phrasing. Importantly, such preprocessing is intended only to stabilize the semantic representation and does not replace psychometric validation.

\paragraph{Degrees of freedom in topic modeling and selection} Topic modeling introduces tunable choices (e.g., clustering hyperparameters and representative selection rules) that can affect the selected items. We therefore emphasize transparent reporting of key settings in our article, and recommend routine stability checks under reasonable perturbations. When the stability is low, researchers may consider retaining more items per factor or incorporating additional constraints (e.g., theory-guided inclusion rules) before proceeding to confirmatory testing.

\paragraph{Short-form length constraints} Very short scales reduce redundancy and can amplify the influence of item idiosyncrasies. However, this also makes it harder to preserve within-factor breadth and to maintain stable structural estimation. Thus, the feasibility of aggressive reduction depends not only on the pipeline but also on the scale’s inherent semantic coherence and the minimum number of indicators required for reliable measurement.

\subsection{Future Directions}
The present work opens several directions for extending and strengthening the semantic - topic modeling-based scale simplification method:
First, building on the initial cross-language and cross-cultural validation provided in this study, future work may pursue more systematic examinations of generalizability across languages and populations. Such efforts could involve broader multilingual benchmarks, explicit translation-based comparisons, and formal measurement invariance analyses to disentangle linguistic, cultural, and semantic sources of variation.

Second, future work can formalize a practical human-machine collaboration workflow for scale construction and simplification. In applied settings, we envision a staged process in which the framework first produces an inspectable semantic organization and candidate representatives (e.g., cluster keywords, exemplar items, and a structured summary for selection), after which substantive experts conduct a targeted review to confirm content coverage, resolve ambiguous boundaries, and ensure alignment with intended construct definitions. When the response data are available, this same process can incorporate response-informed evidence to prioritize candidates with stronger psychometric utility (e.g., higher loadings or information), providing an additional refinement signal alongside semantic representativeness. Framed this way, the framework functions as a transparent front-end that accelerates and documents early decisions, while expert review and downstream validation provide principled refinement and confirmation.

Finally, future work can examine how text-driven selection relates to traditional response-driven reduction (e.g., IRT). Rather than treating these methods as substitutes, a controlled comparison using the same item pool and sample could reveal differences in selection priorities, for example, content representativeness versus information efficiency, and clarify when the two approaches converge, diverge, or can be combined.

\newpage
\section{The One-Click Psychological Scale Simplification Tool}
To facilitate adoption of the proposed framework by researchers without programming experience, we implemented the topic modeling-based simplification method as an interactive, one-click software tool with a graphical user interface. 
The tool provides an end-to-end implementation of the response-free semantic pipeline described in the preceding sections, enabling psychologists to explore semantic structure, identify representative items, and generate interpretable visualizations directly from item texts, without writing code or configuring complex pipelines. In this way, the tool thus serves as a practical bridge between methodological innovation and routine psychometric practice.

\paragraph{Functionality} Specifically, the software implements the proposed framework as a set of tightly integrated functional modules: (a) direct input and preprocessing of questionnaire item texts, (b) semantic encoding and response-free structure discovery via clustering and topic modeling, (c) interpretable topic labeling and representative item selection based on probabilistic topic membership, and (d) visualization of semantic organization in a low-dimensional embedding space.

\begin{figure}[ht]
    \caption{One Click Simplification Tool Interface (Idle, Steps 1 and 2)}
    \includegraphics[width=1\linewidth]{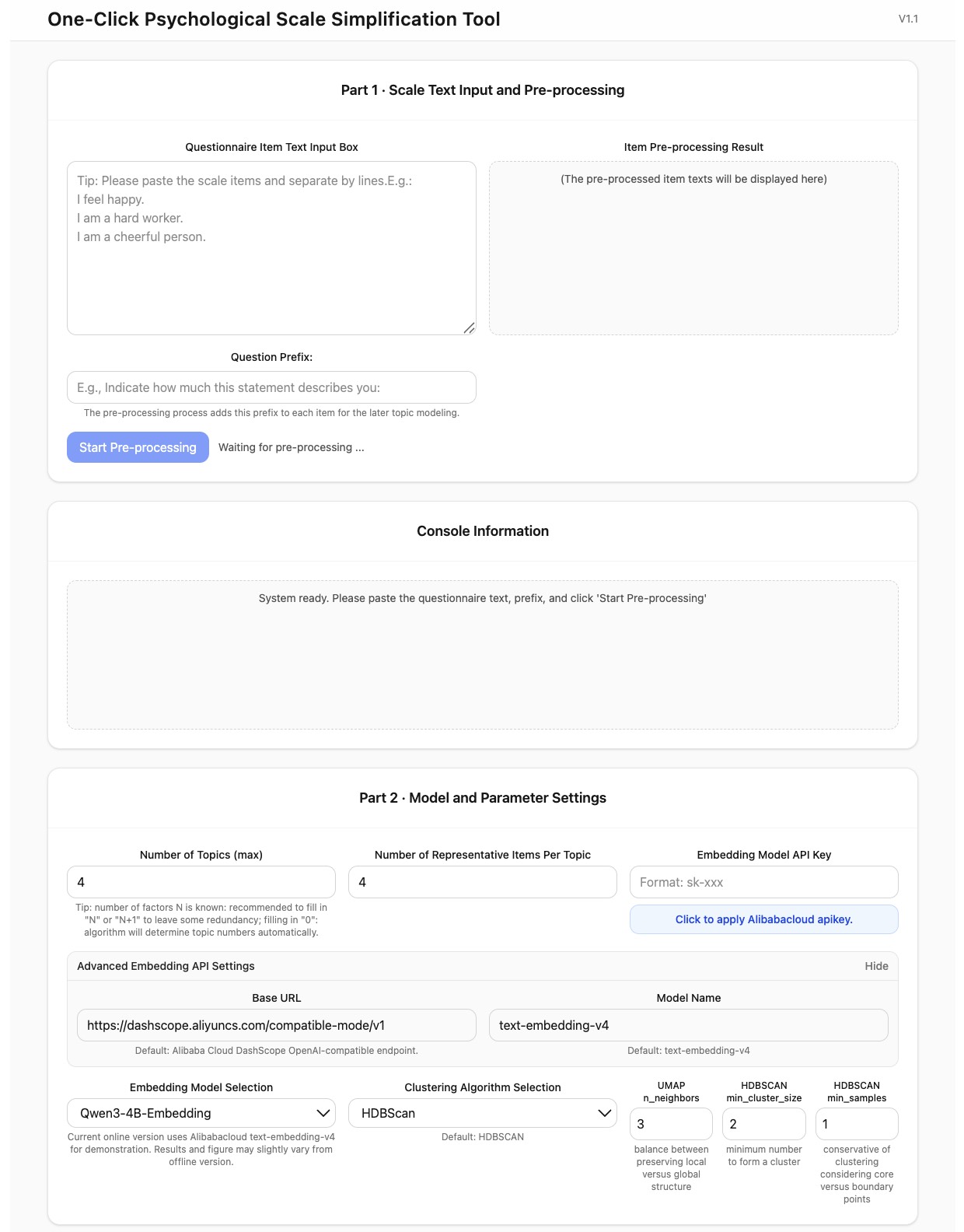}
    \label{fig:tool-1}
\end{figure}

\begin{figure}[ht]
    \caption{One Click Simplification Tool Interface (Idle, Steps 3 and 4)}
    \includegraphics[width=1\linewidth]{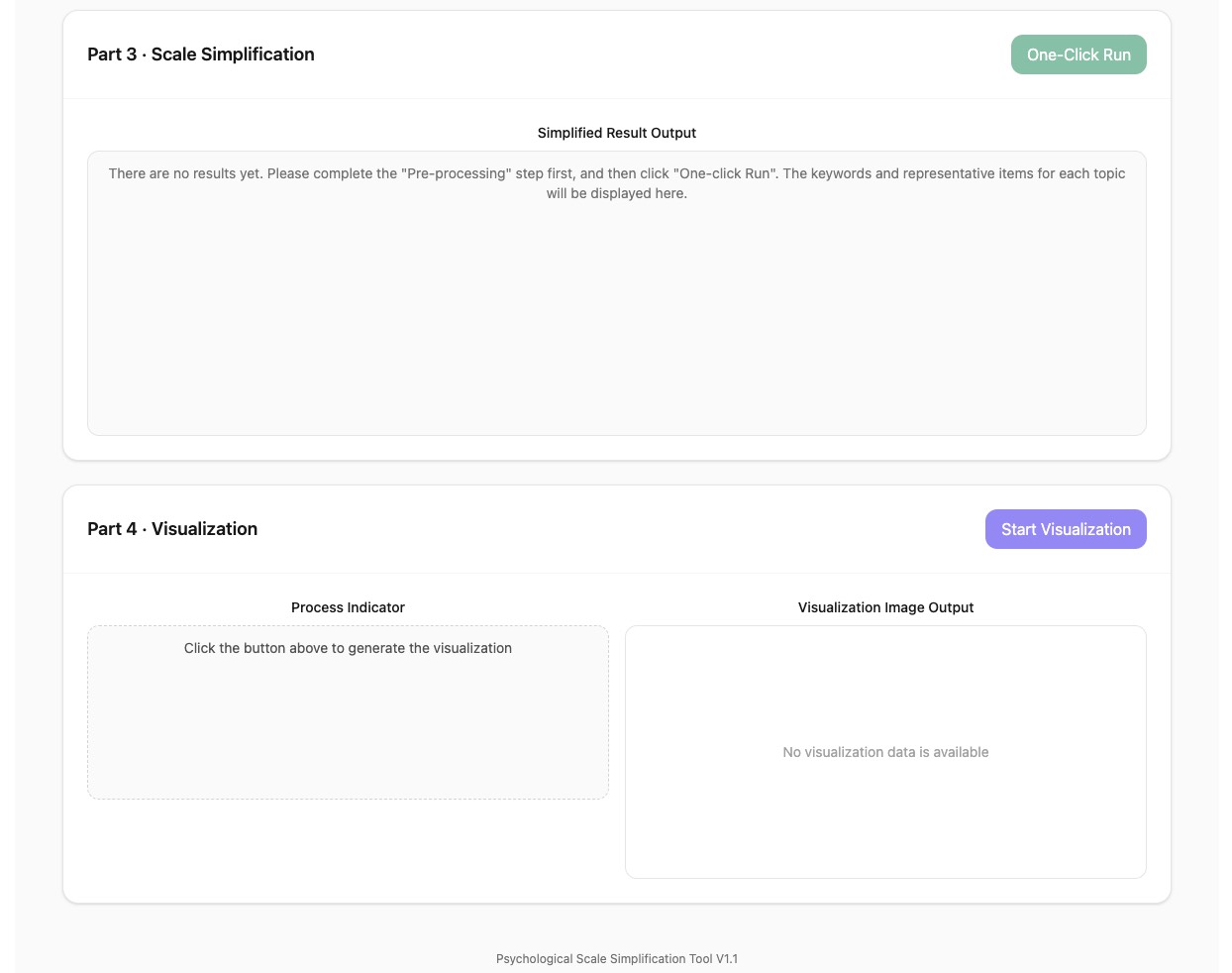}
    \label{fig:tool-2}
\end{figure}

\begin{figure}[ht]
    \caption{One Click Simplification Tool Interface (Working on DASS simplification, Steps 1 and 2)}
    \includegraphics[width=1\linewidth]{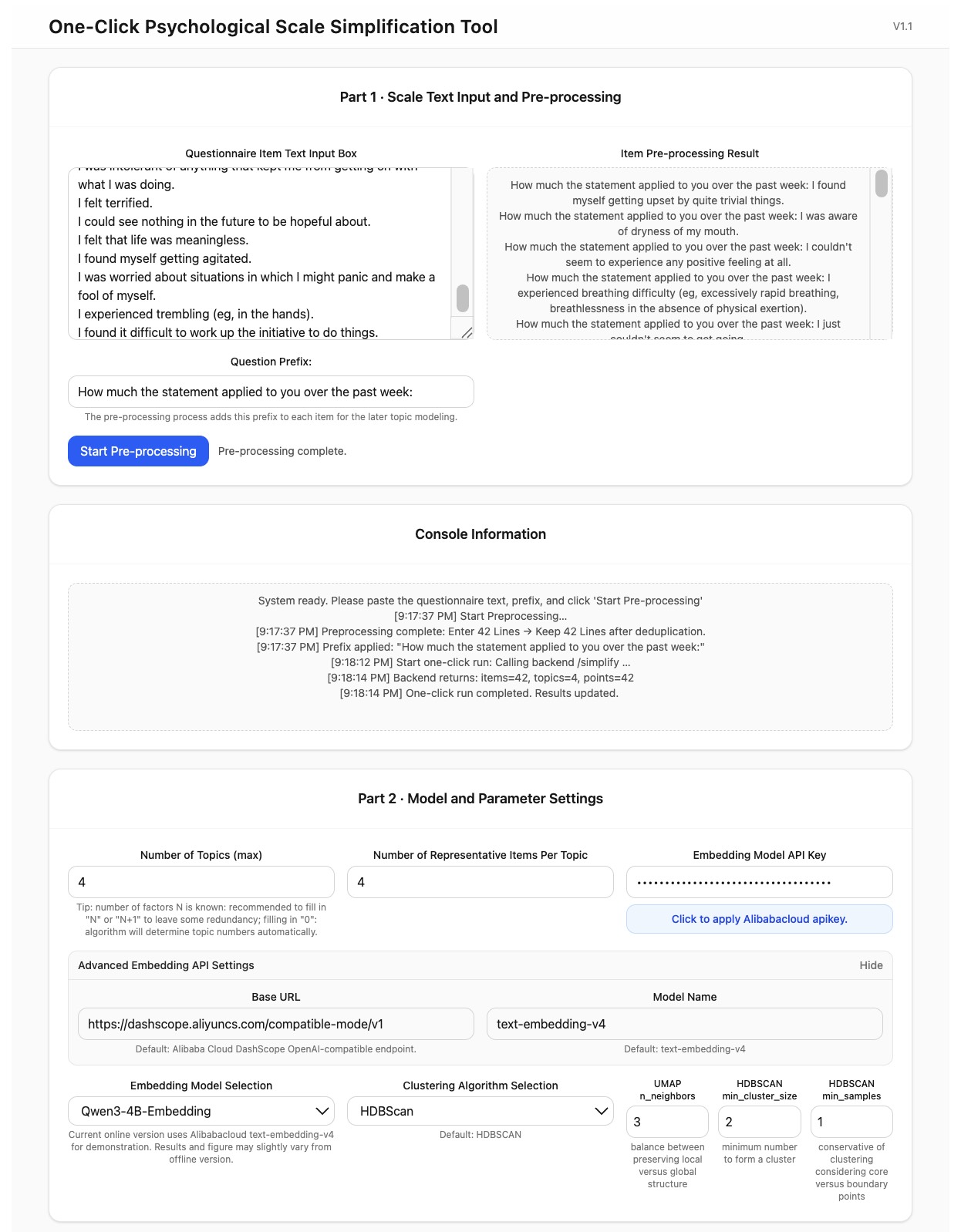}
    \label{fig:tool-3}
\end{figure}

\begin{figure}[ht]
    \caption{One Click Simplification Tool Interface (Working on DASS simplification, Steps 3 and 4)}
    \includegraphics[width=1\linewidth]{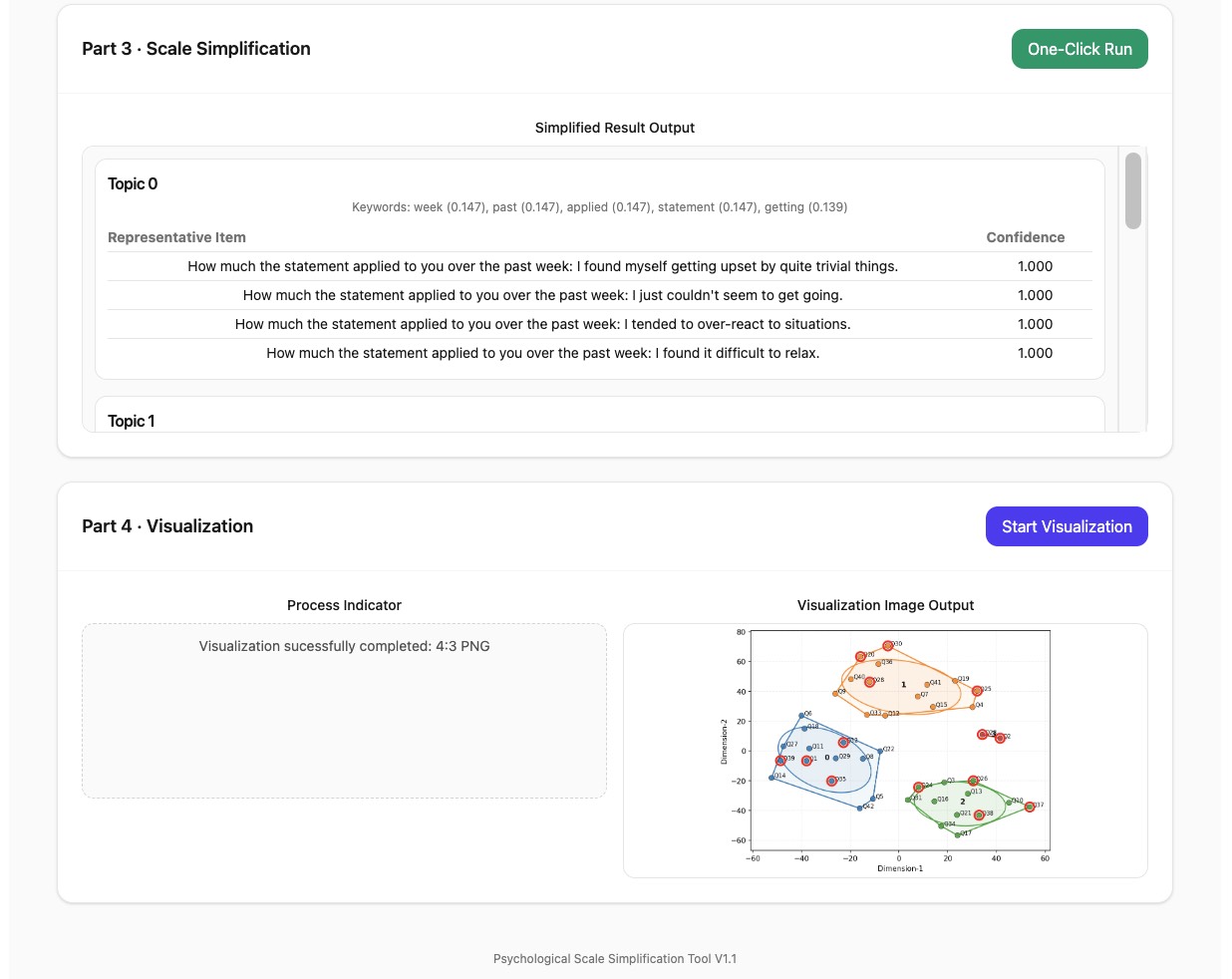}
    \label{fig:tool-4}
\end{figure}

\paragraph{Interface Design and User Workflow} The graphical interface is organized into four sequential panels that correspond to the main stages of the framework and guide users through a transparent, step-by-step workflow (Figures \ref{fig:tool-1},\ref{fig:tool-2},\ref{fig:tool-3},\ref{fig:tool-4}).

\subparagraph{Part 1: Scale Text Input and Pre-processing}
Users paste questionnaire items directly into the input box and add the corresponding question prefix. After preprocessing, the processed item texts are displayed to allow inspection before further analysis.
Note again that if the scale contains reverse-worded items, currently users are required to manually rewrite these items into their counterparts forms prior to input. An example of this rewriting procedure using the IPIP scale and the example rephrasing prompt are provided in Appendices A and B, respectively.

\subparagraph{Part 2: Model and Parameter Settings}
High-level parameters, such as the maximum number of topics and the number of representative items per topic, are exposed with recommended defaults. Advanced parameters related to dimensionality reduction and clustering are visually separated and accompanied by guidance encouraging users to retain default values unless they have specific methodological reasons to modify them.

\subparagraph{Part 3: One-Click Scale Simplification}
With pre-processing completed, users can execute the full pipeline with a single action. The output panel presents the discovered semantic topics, the associated keywords, and the selected representative items for each topic.

\subparagraph{Part 4: Visualization}
A visualization panel generates a two-dimensional embedding map with each topic-cluster coloring and item annotations. This visualization supports qualitative inspection and explanation of the semantic structure and item-selection results, helping users understand why particular items were grouped and selected.

\paragraph{Notes and Software Availability} Importantly, the software is intended as an instantiation of the proposed simplification framework rather than a replacement for response-based psychometric validation, nor as a full-featured psychometric analysis platform. It does not perform confirmatory factor analysis, reliability estimation, or model comparison. Instead, it produces an explicit and inspectable semantic pre-structure that can inform subsequent psychometric evaluation and decision-making.

The tool with source code, documentation, and example materials are available at \url{https://github.com/bowang-rw-02/sem-scale} to support transparency, reproducibility, and reuse. A standalone Python code implementation is also provided for researchers who wish to incorporate the method into custom analysis pipelines or conduct large-scale experiments programmatically.

\newpage
\section{Conclusion}
In this paper, we introduced a semantic - topic modeling framework for psychological scale simplification that connects item-level semantic structure to standard measurement practice. Using topic modeling, the framework organizes item content and selects representative candidates prior to analyzing response data, yielding a transparent and reproducible front-end for short-form construction. Across three case studies, the resulting short forms showed acceptable psychometric performance under conventional evaluations (e.g., CFA, reliability, and cross-form correspondence), supporting the feasibility of the proposed method.

More broadly, these results also draw a pragmatic view of computational text analysis as a complement to psychometrics: semantic modeling can reduce the search space of candidate items and clarify content coverage, but does not replace measurement evaluation. To facilitate adoption, we also provide a one-click tool that operationalizes the full pipeline and makes key intermediate outputs inspectable, thereby lowering the barrier to transparent scale reduction in applied settings. Future work will further examine the framework's generalizability across languages and populations and clarify how the semantic topic modeling-based method complements traditional response-driven simplification methods.

\newpage
\section*{Declarations}
\paragraph{Funding}
This study was funded by the Shenzhen Natural Science Foundation (No. JCYJ20241202124000001), Shenzhen Science and Technology Program (Grant No. KCXFZ20230731093600002), Natural Science Foundation of Guangdong Province (Grant No.  2025A1515010223), and Shenzhen Philosophy and Social Science Foundation (Grant No. SZ2025B016) and Shenzhen Key Laboratory of Philosophy and Social Sciences on AI  and Mental Health (Grant No. 202601).

\paragraph{Competing interests}
The authors declare that they have no competing interests.

\paragraph{Ethics approval}
This study did not involve the collection of new data from human participants. 
All analyses were conducted using previously collected or publicly available datasets. 
According to institutional research ethics guidelines, studies based solely on publicly available or previously collected data are exempt from formal ethical approval.

\paragraph{Consent to participate}
Not applicable.

\paragraph{Consent for publication}
Not applicable.

\paragraph{Availability of data and materials}
The datasets analyzed in this study are publicly available or previously published sources referenced in the manuscript. 

DASS: 
\url{https://openpsychometrics.org/_rawdata/DASS_data_21.02.19.zip}

IPIP:
\url{https://openpsychometrics.org/_rawdata/IPIP-FFM-data-8Nov2018.zip}

EPOCH-CN:
\url{https://osf.io/9ts76/files}

\paragraph{Code availability}
All analysis code and research materials necessary to reproduce the proposed framework are available at: \url{https://github.com/bowang-rw-02/sem-scale}.


\section*{Open Practices Statement}
Analysis code and research materials necessary to reproduce the proposed framework are publicly available at \url{https://github.com/bowang-rw-02/sem-scale}. 
The datasets analyzed in this study are publicly available or previously published sources cited in the manuscript. 
The present study was not preregistered.

\printbibliography

\appendix

\section{Scale Full Question List}
\label{app:qlist}

\subsection{DASS}
Question No./	Factor/	Item text	\\
Q1	S	I found myself getting upset by quite trivial things.	\\
Q2	A	I was aware of dryness of my mouth.	\\
Q3	D	I couldn't seem to experience any positive feeling at all.	\\
Q4	A	I experienced breathing difficulty (eg, excessively rapid breathing, breathlessness in the absence of physical exertion).	\\
Q5	D	I just couldn't seem to get going.	\\
Q6	S	I tended to over-react to situations.	\\
Q7	A	I had a feeling of shakiness (eg, legs going to give way).	\\
Q8	S	I found it difficult to relax.	\\
Q9	A	I found myself in situations that made me so anxious I was most relieved when they ended.	\\
Q10	D	I felt that I had nothing to look forward to.	\\
Q11	S	I found myself getting upset rather easily.	\\
Q12	S	I felt that I was using a lot of nervous energy.	\\
Q13	D	I felt sad and depressed.	\\
Q14	S	I found myself getting impatient when I was delayed in any way (eg, elevators, traffic lights, being kept waiting).	\\
Q15	A	I had a feeling of faintness.	\\
Q16	D	I felt that I had lost interest in just about everything.	\\
Q17	D	I felt I wasn't worth much as a person.	\\
Q18	S	I felt that I was rather touchy.	\\
Q19	A	I perspired noticeably (eg, hands sweaty) in the absence of high temperatures or physical exertion.	\\
Q20	A	I felt scared without any good reason.	\\
Q21	D	I felt that life wasn't worthwhile.	\\
Q22	S	I found it hard to wind down.	\\
Q23	A	I had difficulty in swallowing.	\\
Q24	D	I couldn't seem to get any enjoyment out of the things I did.	\\
Q25	A	I was aware of the action of my heart in the absence of physical exertion (eg, sense of heart rate increase, heart missing a beat).	\\
Q26	D	I felt down-hearted and blue.	\\
Q27	S	I found that I was very irritable.	\\
Q28	A	I felt I was close to panic.	\\
Q29	S	I found it hard to calm down after something upset me.	\\
Q30	A	I feared that I would be "thrown" by some trivial but unfamiliar task.	\\
Q31	D	I was unable to become enthusiastic about anything.	\\
Q32	S	I found it difficult to tolerate interruptions to what I was doing.	\\
Q33	S	I was in a state of nervous tension.	\\
Q34	D	I felt I was pretty worthless.	\\
Q35	S	I was intolerant of anything that kept me from getting on with what I was doing.	\\
Q36	A	I felt terrified.	\\
Q37	D	I could see nothing in the future to be hopeful about.	\\
Q38	D	I felt that life was meaningless.	\\
Q39	S	I found myself getting agitated.	\\
Q40	A	I was worried about situations in which I might panic and make a fool of myself.	\\
Q41	A	I experienced trembling (eg, in the hands).	\\
Q42	D	I found it difficult to work up the initiative to do things.	\\

\subsection{IPIP}
Qid	Reversed/	Original factor-id/	Original item text/	Processed item text (reversed item was rewritten)	\\
1		EXT1	I am the life of the party.	I am the life of the party.	\\
2	\textit{(R)}	EXT2	I don't talk a lot.	I talk a lot.	\\
3		EXT3	I feel comfortable around people.	I feel comfortable around people.	\\
4	\textit{(R)}	EXT4	I keep in the background.	I put myself forward.	\\
5		EXT5	I start conversations.	I start conversations.	\\
6	\textit{(R)}	EXT6	I have little to say.	I have a lot to say.	\\
7		EXT7	I talk to a lot of different people at parties.	I talk to a lot of different people at parties.	\\
8	\textit{(R)}	EXT8	I don't like to draw attention to myself.	I like to draw attention to myself.	\\
9		EXT9	I don't mind being the center of attention.	I don't mind being the center of attention.	\\
10	\textit{(R)}	EXT10	I am quiet around strangers.	I am talkative around strangers.	\\
11	\textit{(R)}	EST1	I get stressed out easily.	I stay calm under pressure.	\\
12		EST2	I am relaxed most of the time.	I am relaxed most of the time.	\\
13	\textit{(R)}	EST3	I worry about things.	I am generally free from worry.	\\
14		EST4	I seldom feel blue.	I seldom feel blue.	\\
15	\textit{(R)}	EST5	I am easily disturbed.	I remain composed and unbothered.	\\
16	\textit{(R)}	EST6	I get upset easily.	I stay calm and even-tempered.	\\
17	\textit{(R)}	EST7	I change my mood a lot.	I stay in a steady mood.	\\
18	\textit{(R)}	EST8	I have frequent mood swings.	I remain emotionally stable.	\\
19	\textit{(R)}	EST9	I get irritated easily.	I stay patient and calm.	\\
20	\textit{(R)}	EST10	I often feel blue.	I feel cheerful and upbeat.	\\
21	\textit{(R)}	AGR1	I feel little concern for others.	I feel concern and empathy for others.	\\
22		AGR2	I am interested in people.	I am interested in people.	\\
23	\textit{(R)}	AGR3	I insult people.	I treat people with respect.	\\
24		AGR4	I sympathize with others' feelings.	I sympathize with others' feelings.	\\
25	\textit{(R)}	AGR5	I am not interested in other people's problems.	I care about other people's problems.	\\
26		AGR6	I have a soft heart.	I have a soft heart.	\\
27	\textit{(R)}	AGR7	I am not really interested in others.	I am genuinely interested in others.	\\
28		AGR8	I take time out for others.	I take time out for others.	\\
29		AGR9	I feel others' emotions.	I feel others' emotions.	\\
30		AGR10	I make people feel at ease.	I make people feel at ease.	\\
31		CSN1	I am always prepared.	I am always prepared.	\\
32	\textit{(R)}	CSN2	I leave my belongings around.	I keep my belongings organized.	\\
33		CSN3	I pay attention to details.	I pay attention to details.	\\
34	\textit{(R)}	CSN4	I make a mess of things.	I keep things neat and in order.	\\
35		CSN5	I get chores done right away.	I get chores done right away.	\\
36	\textit{(R)}	CSN6	I often forget to put things back in their proper place.	I put things back in their proper place.	\\
37		CSN7	I like order.	I like order.	\\
38	\textit{(R)}	CSN8	I shirk my duties.	I take responsibility for my duties.	\\
39		CSN9	I follow a schedule.	I follow a schedule.	\\
40		CSN10	I am exacting in my work.	I am exacting in my work.	\\
41		OPN1	I have a rich vocabulary.	I have a rich vocabulary.	\\
42	\textit{(R)}	OPN2	I have difficulty understanding abstract ideas.	I easily understand abstract ideas.	\\
43		OPN3	I have a vivid imagination.	I have a vivid imagination.	\\
44	\textit{(R)}	OPN4	I am not interested in abstract ideas.	I am interested in abstract ideas.	\\
45		OPN5	I have excellent ideas.	I have excellent ideas.	\\
46	\textit{(R)}	OPN6	I do not have a good imagination.	I have a good imagination.	\\
47		OPN7	I am quick to understand things.	I am quick to understand things.	\\
48		OPN8	I use difficult words.	I use difficult words.	\\
49		OPN9	I spend time reflecting on things.	I spend time reflecting on things.	\\
50		OPN10	I am full of ideas.	I am full of ideas.	\\

\subsection{EPOCH-CN and original EPOCH}
Qid/	Original factor-id/	Item text in Chinese/	Corresponding English item text	\\
1	C1	\begin{CJK}{UTF8}{gbsn}当有好事发生在我身上的时候，我有喜欢的人可以去分享。\end{CJK}	When something good happens to me, I have people who I like to share the good news with.	\\
2	P1	\begin{CJK}{UTF8}{gbsn}只要开始做一件事情，我就会完成它。\end{CJK}	I finish whatever I begin.	\\
3	O1	\begin{CJK}{UTF8}{gbsn}我对自己的未来充满乐观。\end{CJK}	I am optimistic about my future	\\
4	H1	\begin{CJK}{UTF8}{gbsn}我很快乐。\end{CJK}	I feel happy.	\\
5	E1	\begin{CJK}{UTF8}{gbsn}当我做一项活动的时候，我非常乐在其中以至于忘了时间。\end{CJK}	When I do an activity, I enjoy it so much that I lose track of time.	\\
6	H2	\begin{CJK}{UTF8}{gbsn}我有很多的欢乐。\end{CJK}	I have a lot of fun.	\\
7	E2	\begin{CJK}{UTF8}{gbsn}我会全神贯注于我正在做的事。\end{CJK}	I get completely absorbed in what I am doing.	\\
8	H3	\begin{CJK}{UTF8}{gbsn}我热爱生活。\end{CJK}	I love life.	\\
9	P2	\begin{CJK}{UTF8}{gbsn}我会一直坚持做作业直到完成为止。\end{CJK}	I keep at my schoolwork until I am done with it.	\\
10	C2	\begin{CJK}{UTF8}{gbsn}当我遇到问题时，总有人支持我。\end{CJK}	When I have a problem, I have someone who will be there for me.	\\
11	E3	\begin{CJK}{UTF8}{gbsn}我会专心致志地投入到当下的活动以至于忘记了其他事情。\end{CJK}	I get so involved in activities that I forget about everything else.	\\
12	E4	\begin{CJK}{UTF8}{gbsn}当我学习新东西时，我是那么地投入以至于忘记时间的流逝。\end{CJK}	When I am learning something new, I lose track of how much time has passed.	\\
13	O2	\begin{CJK}{UTF8}{gbsn}在不明确的情况下，我期待最好的结果。\end{CJK}	In uncertain times, I expect the best.	\\
14	C3	\begin{CJK}{UTF8}{gbsn}我的生命中有一些人真心地关心我。\end{CJK}	There are people in my life who really care about me.	\\
15	O3	\begin{CJK}{UTF8}{gbsn}我相信好事会发生在我身上。\end{CJK}	I think good things are going to happen to me.	\\
16	C4	\begin{CJK}{UTF8}{gbsn}我有我真正关心的朋友。\end{CJK}	I have friends that I really care about.	\\
17	P3	\begin{CJK}{UTF8}{gbsn}一旦我计划了要做某事，我就会按计划进行。\end{CJK}	Once I make a plan to get something done, I stick to it.	\\
18	O4	\begin{CJK}{UTF8}{gbsn}我相信事情总会好起来，不管它看起来有多困难。\end{CJK}	I believe that things will work out, no matter how difficult they seem.	\\
19	P4	\begin{CJK}{UTF8}{gbsn}我是一个勤奋的人。\end{CJK}	I am a hard worker.	\\
20	H4	\begin{CJK}{UTF8}{gbsn}我是一个快乐的人。\end{CJK}	I am a cheerful person.	\\

\section{Rephrasing Procedure and Prompt for Reverse-Keyed Items}

Before semantic encoding, reverse-keyed items were rephrased into positive-keyed statements to align the semantic direction of item texts. Although such items are handled through reverse scoring in psychometric analysis, text encoders may treat negation and wording polarity (such as ``not'', ``don't'', or ``never'') as salient semantic features, potentially separating items that measure the same construct but are expressed in opposite directions. This preprocessing step was therefore used to reduce negation-related artifacts and to make the resulting embeddings, clusters, and topics more reflective of the underlying constructs rather than item wording direction.

In our study, the prompt used to rephrase the IPIP reverse-keyed items is provided below. The rephrasing was conducted using ChatGPT (GPT-5.1 Instant).

\begin{quote}
	\small \itshape
	You are a psychologist with expertise in psychometrics and questionnaire item construction. I will provide a list of reverse-keyed questionnaire items. Please rewrite each item into a positive-keyed version that measures the same psychological construct but in the opposite direction.
	
	Please follow these principles:
	
	1. Preserve the original meaning and psychological construct as closely as possible.
	
	2. Make only minimal wording changes when possible.
	
	3. Keep the wording concise, natural, and suitable for self-report questionnaire items.
	
	4. Avoid adding new examples, contexts, or psychological meanings.
	
	5. Avoid double negatives and overly strong or ambiguous wording.
	
	Example:
	
	Original item: ``I don’t talk a lot.''
	Rewritten item: ``I talk a lot.''
	
	Original item: ``I worry about things.''
	Rewritten item: “I am generally free from worry.''
	
	Now rewrite the following reverse-keyed items and output a two-column table with ``Original item'' and ``Rewritten item'':
	
	[ITEM LIST]
\end{quote}

\end{document}